\documentclass{article}

\usepackage{microtype}
\usepackage{graphicx}
\usepackage{subfigure}
\usepackage{booktabs} % for professional tables

\usepackage{hyperref}

\usepackage[accepted]{icml2019}

\usepackage{amsmath,amsthm,amssymb,bm}
\usepackage{xcolor}
\usepackage[toc,page]{appendix}

\newcommand{\R}{\mathbb{R}}

\newcommand{\E}{\mathbb{E}}

\DeclareMathOperator*{\cov}{cov}
\DeclareMathOperator*{\var}{var}

\newtheorem{proposition}{Proposition}
\newtheorem{corollary}{Corollary}

\usepackage{color}

\usepackage{xcolor}
\definecolor{dark-red}{rgb}{0.4,0.15,0.15}
\definecolor{dark-blue}{rgb}{0.15,0.15,0.4}
\definecolor{medium-blue}{rgb}{0,0,0.5}
\hypersetup{
   colorlinks, linkcolor={dark-blue},
   citecolor={dark-blue}, urlcolor={medium-blue}
}

\icmltitlerunning{Randomly Projected Additive Gaussian Processes}

\begin{document}

\twocolumn[
\icmltitle{Randomly Projected Additive Gaussian Processes for Regression}
\icmlsetsymbol{equal}{*}

\begin{icmlauthorlist}
\icmlauthor{Ian A. Delbridge}{corn}
\icmlauthor{David S. Bindel}{corn}
\icmlauthor{Andrew Gordon Wilson}{corn,nyu}\\
$^1$Cornell University, $^2$New York University
\end{icmlauthorlist}

\icmlaffiliation{corn}{Department of Computer Science, Cornell University, Ithaca, NY USA}
\icmlaffiliation{nyu}{Courant Institute, New York University, New York, NY USA}

\icmlcorrespondingauthor{Ian A. Delbridge}{iad35@cornell.edu}
\icmlcorrespondingauthor{Andrew G. Wilson}{andrewgw@cims.nyu.edu}

% You may provide any keywords that you
% find helpful for describing your paper; these are used to populate
% the "keywords" metadata in the PDF but will not be shown in the document
\icmlkeywords{Machine Learning, Gaussian Processes, Random Projection, Additive}

\vskip 0.3in
]

\begin{abstract} 

Gaussian processes (GPs) provide flexible distributions over functions, with inductive biases controlled by a kernel. However, in many applications 
Gaussian processes can struggle with even moderate input 
dimensionality. Learning a low dimensional projection can help alleviate this curse of dimensionality, but introduces many trainable hyperparameters, which can 
be cumbersome, especially in the small data regime. We use additive sums of kernels for GP regression, where each kernel operates on a different random projection of its inputs. 
Surprisingly, 
we find that as the number of random projections increases, the predictive performance of this approach quickly converges to the performance of a kernel operating on the original full 
dimensional inputs, over a wide range of data sets, \emph{even if we are projecting into a single dimension}. As a consequence, many problems can remarkably be reduced to one dimensional input spaces, without learning a transformation. We prove this convergence and its rate, and additionally propose a deterministic approach 
that converges more quickly than purely random projections. Moreover, we demonstrate our approach can achieve faster inference and improved predictive accuracy for high-dimensional inputs compared to kernels in the original input space. 

\end{abstract}

\section{Introduction}
\label{Introduction}

Gaussian processes (GPs) are flexible Bayesian non-parametric models with well-calibrated predictive uncertainties. Gaussian processes can also naturally encode inductive biases, such as smoothness or periodicity, through a choice of kernel function \citep{rasmussen2006gaussian}. Gaussian processes have been especially impactful in the small-data regime, where careful uncertainty representation is particularly crucial, strong priors provide useful biases where learning is difficult, and exact inference is tractable. Additionally, Gaussian processes have been most successfully applied to low-dimensional input (predictor) spaces, such as time series, and spatiotemporal regression problems \citep[e.g.,][]{wilson2013gaussian, duvenaud2014automatic, herlands2018change}. In these settings, canonical kernels --- such as the RBF or Mat\'ern kernels --- provide reasonable distance metrics over pairs of data instances; for example, if we are modelling CO$_2$ concentrations indexed by time, then CO$_2$ levels at times which are close together in $\ell_2$ or $\ell_1$ distance will be treated as highly correlated under these kernels.

For higher dimensional problems, these standard distance metrics become less compelling. For example, with an RBF kernel, the fraction of data space with high covariance with a given point decreases exponentially with dimension. Additionally, in many online settings where Gaussian processes are used as regression models, such as Bayesian optimization, there is exponential regret with dimensionality \citep{Srinivas2019Gaussian, bull2011convergence}. Furthermore, scalable Gaussian processes which have a high degree of accuracy often only apply for up to a few input dimensions \citep[e.g.,][]{wilson2015kernel, gilboa2013scaling}.

To help circumvent such issues, there are two popular approaches. The first approach is to \emph{learn} a projection into a lower dimensional space, such as through deep kernel learning \citep{wilson2016deep}. While such approaches are highly flexible, they introduce many hyperparameters to train, which can be burdensome and impractical in the small data regime. Alternatively, additive Gaussian processes \citep{duvenaud2014automatic, kandasamy2015high, hastie1986} instead consider a sum of kernels, with each kernel operating on subsets of the input dimensions. This structure can both help reduce the effective dimensionality of the problem, and provide a useful inductive bias with compelling sample complexity \citep{stone1985additive}. However, while assuming a fully additive decomposition of an untransformed space can provide a useful inductive bias for many real data sets, it is often too restrictive \citep{li2016high}. Moreover, methods for learning additive structure, as with standard projection approaches, are either computationally expensive or require learning a large number of parameters, which may overfit or hurt uncertainty estimation. 

In this work, we show how to dramatically reduce the input dimensionality of a given problem, while retaining or even improving predictive accuracy, without having to learn projections. Specifically, our contributions are as follows: 
\begin{itemize}
    \item We propose a novel learning-free algorithm for constructing additive GPs based on sequences of multiple random projections (RPA-GP). This results in a Turning Band style \citep{matheron197.intrinsic} approximation to a high-dimensional kernel.
    \item  We prove that RPA-GP converges to a full-degree inverse multiquadratic kernel as the number of projections increase at a rate of $\mathcal{O}(J^{-1/2})$ where $J$ is the number of projections.
    \item We propose a deterministic algorithm (DPA-GP) to minimize projection redundancy and achieve faster convergence to the limiting kernel.
    \item We demonstrate the surprising result that RPA-GP and DPA-GP converge very quickly to the regression accuracy of a kernel operating on the full dimensional inputs, over a wide range of regression problems, \emph{even for projections into a single input dimension}. 

    \item We show in a large empirical study that fully additive GPs  can also perform competitively with GPs using standard kernels, but are outperformed by DPA-GP with automatic relevance determination on the original input space, particularly on large data sets and high dimensional data sets.
    \item We additionally demonstrate that by exploiting the additive structure of RPA-GP, we alleviate the curse of dimensionality for structured kernel interpolation (SKI) \citep{wilson2015kernel}, enabling linear-time training and constant-time predictions over a wide range of problems, including problems with over $1000$ input dimensions.
   \item We provide GPyTorch code \citep{gardner2018gpytorch}  for all models at 
   \url{https://github.com/idelbrid/Randomly-Projected-Additive-GPs}.
   
\end{itemize} 

The high level idea of random projections to compose additive kernels has been considered in geostatistics under the name the \emph{turning band method} (TBM) \citep{matheron197.intrinsic}, for 2 and 3-dimensional \emph{simulation}. However, the execution and details are very different from what we consider here. Our paper analyzes and demonstrates how learning-free additive projections can be promising for \emph{regression} in high dimensional input spaces.

RPA-GP and DPA-GP are a step towards alleviating the curse of dimensionality for Gaussian processes, while retaining a pleasingly tractable and lightweight representation. We focus our experiments on regression, since regression is the basic foundation for many popular procedures involving Gaussian processes, such as Bayesian optimization \citep{movckus1975bayesian}, and model based reinforcement learning \citep{deisenroth2011pilco, Engel2005reinforcement}, and is in itself a widespread application for Gaussian processes \citep{williams1996gaussian, vanBeers2004kriging}.

\section{Background} \label{background}
We briefly review Gaussian process regression and structured kernel interpolation (SKI) \citep{wilson2015kernel}. For more details on Gaussian processes, we refer the reader to \citet{rasmussen2006gaussian}.

\subsection{Gaussian process regression}
Formally, a Gaussian process $f$ is a stochastic process over an index set $\mathcal{X}$ (typically elements of $\mathcal{X}$ are in $\R^d$) taking on real values. Therefore, it can be interpreted as a prior over functions from $\mathcal{X}$ to $\R$. The process evaluated at any finite collection of points is distributed according to a multivariate normal distribution. That is, for any $\bm x_1, ..., \bm x_n \in \mathcal{X}$, $\bm f = [f(\bm x_1), ..., f(\bm x_n)] \sim \mathcal{N}(\bm m_X, K_{X,X})$. Accordingly, a Gaussian process is fully determined by its prior mean function $m: \mathcal{X} \mapsto \R$ and covariance kernel function $k: \mathcal{X} \times \mathcal{X} \mapsto \R$. The prior mean function is often chosen to be $0$ in the case where we have limited knowledge of $f$. Therefore, a Gaussian process is almost entirely determined by $k$. Standard identities of the multivariate Gaussian distribution can be applied to find the posterior predictive distribution under a Gaussian observation model given data $X,\bm y =\{(\bm x_i, y_i)\}_{i=1}^n$ at points $X_*$ is 
\begin{align*}
    \bm f_*|X,\bm y, X_* &\sim \mathcal{N}(\bar{\bm{f}}_*, \cov(\bm f_*)),
    \end{align*}
where
\begin{align*}
    \bar{\bm{f}}_* &:= K_{X_*,X}(K_{X,X} + \sigma^2 I_n)^{-1} \bm y, \\
    \cov(\bm f_*) &:= K_{X_*, X_*} - K_{X_*, X}(K_{X, X} + \sigma^2 I_n)^{-1} K_{X, X_*}.
\end{align*}

The computational bottleneck in computing the posterior distribution is solving the linear system $(K_{X,X} + \sigma^2 I_n)^{-1} \bm{y}$. Standard approaches use the Cholesky decomposition, which requires $\mathcal{O}(n^3)$ computations.

The log marginal likelihood 
\begin{align*}
    \log p(\bm y | X) = &- \frac{1}{2}\bm y^{\top}(K_{X,X} + \sigma^2 I_n)^{-1} \bm y \\
    &- \frac{1}{2}\log |K_{X,X} +\sigma^2 I_n| - \frac{n}{2}\log 2 \pi
\end{align*}
is used for model comparison and optimization. Typically, one parameterizes the kernel with some number of hyperparameters which are tuned by maximizing the marginal likelihood. This maximization provides automatic regularization because the determinant $|K_{X,X} + \sigma^2 I_n|$ penalizes quickly varying functions. The computational bottleneck in computing the marginal likelihood is the determinant, which has the standard computational cost of $\mathcal{O}(n^3)$ from the Cholesky decomposition.

\subsection{Structured kernel interpolation}\label{section:background_SKI}

Structured kernel interpolation (SKI) \citep{wilson2015kernel} uses an approximation to the kernel $K_{X,X}$ that permits fast matrix-vector multiplications, which are used to compute the log marginal likelihood and predictive distributions. Specifically, let $U$ be a regular grid of inducing points. \citet{wilson2015kernel} let $W$ be a matrix of local cubic interpolation weights from $U$ to $X$ \citep{keys1981cubic}. The SKI kernel approximation of base kernel matrix $K_{X,X}$ is 
\begin{align*}
    K_{X,X} \approx K^{\text{SKI}}_{X,X} := W K_{U,U} W^\top.
\end{align*}
The interpolation matrix $W$ is sparse, having only $4^d$ nonzero elements per row. The matrix $K_{U,U}$ can have Toeplitz (if $d=1$) or Kronecker (if $d>1$) structure, either of which permit fast matrix-vector multiplications with $K_{U,U}$ \citep{saatcci2012scalable} and thus also the approximate $K_{X,X}$, due to the sparse interpolation in SKI. The linear solve $(K^{\text{SKI}}_{X,X} + \sigma^2 I_n)^{-1} \bm y$ can then be efficiently computed using linear conjugate gradients, which proceeds by iterative matrix-vector multiplications of $K^{\text{SKI}}_{X,X} + \sigma^2 I_n$. The log determinant can be computed using stochastic Lanczos quadrature \citep{dong2017scalable}, which similarly only requires iterative matrix-vector multiplications of $K^{\text{SKI}}_{X,X} + \sigma^2 I_n$. 

However, fixing the number of inducing points in each dimension, the size of the grid grows exponentially with dimension. Therefore, inference using SKI is intractable generally for dimension $d>5$. 

\section{Related Work}\label{Related Work}

GPs with kernels that fully decompose additively\footnote{We define the \textit{degree} of a kernel to be the number of dimensions over which it operates. We say a kernel is \textit{additive} simply if it is a sum of lower-degree kernels. Moreover, a GP is additive if its kernel is additive.}, i.e.
\begin{align}
    k(\bm x, \bm x^\prime) = \sum_{i=1}^d k_i(x_{i}, x^\prime_{i}), \label{full decomposition}
\end{align}
for some sub-kernels $\{k_i\}_{i=1}^d$, are considered Generalized Additive Models (GAM) \citep{hastie1986}. We refer to the resulting GP as a GAM GP throughout this paper. Here, we denote the $i$th component of vector $\bm x$ as $x_{i}$ without bold face to indicate that it is a scalar value.

The GAM GP implicitly assumes there are only first-order interactions in the modeled function. This assumption may be reasonable inductive bias in some cases, but it is often too strong \citep{li2016high, duvenaud2011additive}. It is natural, then, to consider additive combinations of sets of features. Unfortunately, the space of subsets of features is a power set and therefore grows exponentially. Therefore, learning additive combinations of kernels on subsets of features is difficult. Extant approaches to learning additive kernel structure can be divided roughly into \textit{enumeration} methods, where sub-kernels consider every combination of feature interactions up to a degree, \textit{search methods}, where possible decompositions are traversed by a search algorithm, and \textit{projection-pursuit} where a projected-additive GP is learned by iteratively optimizing projection directions from regression residuals.

 Hierarchical Kernel Learning \citep{bach2009high} is an enumeration method in which one constructs the sum of kernels in a hull of possible kernels. \citet{duvenaud2011additive} compute the sum of kernels over every possible feature combination in $O(d^2)$ time by using the Newton-Girard formulae. \citet{duvenaud2013structure} define a grammar over kernels and use discrete search to optimize kernel structure. \citet{qamar2014additive} uses a sampling approach to search through additive decompositions. In Bayesian optimization, it is especially beneficial to learn additive structure where no features are overlapped between sub-kernels. \citet{gardner2017discovering} perform MCMC sampling over such kernel structures as a search method. Similarly, \citet{wang2017batched} perform a Gibbs sampling procedure to search over feature partitions.  Enumeration methods inherently incorporate a very large number of sub-kernels, which can be expensive to compute for high dimensions. Search methods, on the other hand, are burdened by searching over a combinatorial space.
 
Projection pursuit, introduced by \citet{friedman1981projection} and adapted to the Gaussian process setting by \citet{saatcci2012scalable, gilboa2013scaling}, is different in that one learns \textit{projected}-additive GPs. That is, the GP is an additive combination of low-dimensional kernels defined on linear projections of data whose directions are sequentially optimized.  If a large number of projections are used, the sequential optimization of directions with respect to the marginal likelihood can be computationally expensive, and the large number of parameters learned by optimization may result in overfitting and poor uncertainty estimation \citep{li2016high}. 

A GP using a single non-additive random projection has been briefly considered with promising preliminary results \citep{wang2016bayesian}. However, we find that such methods can be dramatically improved through sequences of additive random and deterministic projections, and investigate this surprising and practically significant result. Additionally, \citet{guhaniyogi2016compressed} use a GP over random projections of high-dimensional data having low-dimensional manifold structure. However, their work does not explore additive Gaussian processes and also relies on a model average of many GPs to account for variation in the random projections.

Composing 1-dimensional stochastic processes along random directions to approximate higher-dimensional stochastic processes has been used in the geostatistics community under the name the ``turning bands method'' and has since been studied in detail for simulation of 2 to 3-dimensional processes  \citep{matheron197.intrinsic,mantoglou1982turning,mantoglou1987digital,lantuejoul2013geostatistical}. Work has been devoted to describing the 1-dimensional covariances associated with common covariances \citep{christakos1987stochastic,gneiting1998closed}, quantifying the approximation error \citep{mantoglou1987digital}, and even choosing well-spread directions \citep{freulon1993revisiting,lantuejoul2013geostatistical}. Yet, this direction of work has not been explored for higher dimensional GPs, nor for Gaussian process regression. 

\section{Randomly Projected-Additive GPs} \label{RPGP}

Rather than directly learning additive structure, we project data onto randomly drawn directions and impose additive structure on a GP defined over the projections. As a result, we bypass the need to search over or enumerate all possible sub-kernels, and the burden of training many hyperparameters in a learned projection.

Formally, let $n$ be the number of data points, $d$ be number of dimensions, and $J$ be the number sub-kernels. Denoting the degree of kernel $j$ as $D_j$, we define the randomly projected additive kernel as
\begin{gather}
    k_{rp}(\bm x, \bm x^\prime) = \sum_{j=1}^J \alpha_j k_{j}(P^{(j)}\bm x, P^{(j)}\bm x^\prime), \\
    \forall j \in [J], \hspace{2mm} P^{(j)} \in \R^{D_j \times d}, \\
    P^{(j)}_{r,c} \sim \mathcal{N}\left(0, \frac{1}{D_j}\right) \hspace{2mm} \forall r\in [D_j], c \in [d]. \label{RP sample}
\end{gather}

We refer to a GP with covariance kernel $k_{rp}$ as a randomly-projected additive GP (RPA-GP). Matrices $\{P^{(j)}\}_{j=1}^J$ define the directions of the projections. The parameters $\{\alpha_j\}_{j=1}^J$ determine the amount of variance each sub-kernel contributes and may be either learned or set as a constant value $1/J$.

If the sub-kernel degrees $D_j$ are large enough relative to sample size $n$, the Johnson-Lindenstrauss Lemma guarantees that the $\ell_2$ distances between points are approximately preserved with high probability \citep{sarlos2006improved}. Alternatively, we have a similar guarantee if data lie on a low-dimensional manifold \citep{baraniuk2009random}. Therefore, if we use RBF sub-kernels, each sub-kernel is a good approximation of the high-dimensional RBF kernel. Moreover, having multiple random projections increases the likelihood of drawing a good random projection but does not increase kernel dimensionality, similar to the method presented in \citet{ahmed2004multiple}. 
However, if the sub-kernel degrees are small, we have sufficient flexibility given enough projections. RPA-GP forms a distribution over linear combinations of \textit{ridge functions}, which are defined as functions that are invariant in all but 1 direction. Since linear combinations of ridge functions are dense in the set of continuous functions, we are able to approximate any continuous function arbitrarily well given a rich enough set of directions \citep{cheney2009course}.

\subsection{The expected kernel}\label{section:expected_kernel}

\newcommand{\figwidth}{97pt}
\newcommand{\figheight}{70pt}
\begin{figure}
\centering     %%% not \center
\includegraphics[width=\linewidth]{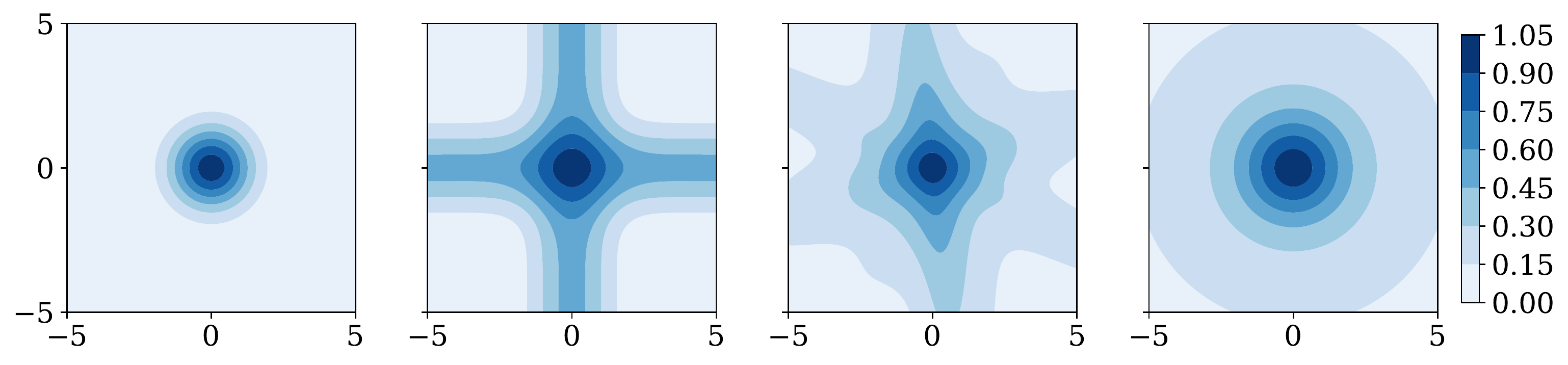}
\caption{Contour plots of 2-dimensional kernels. From left to right: RBF, GAM RBF, RPA-GP with 16 projections, and DPA-GP with 16 projections. With enough additive projections, we attain approximately spherical covariance, and choosing well-placed directions facilitates convergence.}
\label{fig:covariance-visualization}
\end{figure}

We now analyze projected-additive kernels by studying the functional form of the covariance in comparison to the RBF kernel. We limit our analysis to the most challenging case of one-dimensional additive projections, i.e. when each matrix $P^{(j)}$ is a vector $\bm \eta_j$, though analysis is similar for higher dimensional projections. Further, we assume unit length-scale, which can always be achieved by appropriate scaling of data. For brevity, we defer proofs to the appendix.

Clearly, an additive (GAM) covariance kernel does not decay to zero as the $\ell_2$ distance between points goes to infinity. For example, a GAM kernel with RBF additive components is lower bounded by $\frac{d-1}{d}$ along each axis. Conversely, in expectation, the covariance of a randomly projected additive kernel with RBF components decays to zero in any given direction; if $\bm \eta_j$ are drawn from an isotropic distribution, an additive randomly projected kernel converges to a high-dimensional kernel as $J \rightarrow \infty$. This is made formal in the following proposition.

%%% Keeping for now. The previous text in this section %%%
% Consider for example the full-degree RBF kernel and the additive projected RBF kernel
% \begin{gather*}
%     k(\bm \tau) = \exp\left(-\frac{1}{2}||\bm \tau||_2^2\right) \\ \tilde k(\bm \tau) = \frac{1}{J}\sum_{j=1}^J \exp\left(-\frac{1}{2} (\bm \eta_j^\top \bm \tau)^2\right)
% \end{gather*}
% respectively. A special case of $\tilde k$ is when $J=d$ and directions $\{\bm \eta_j\}_{j=1}^J$ form the canonical basis for $\R^d$. In this case, the covariance is a GAM kernel. 

% First, observe that, by the Cauchy-Schwarz inequality, $\tilde k(\bm \tau) \ge k(\bm \tau)$ for all $\bm \tau$ if directions have unit length. Thus, projected additive RBF kernels assign higher covariance compared to the full-degree RBF kernel. At a given distance from the origin, the difference is greatest when $\bm \tau$ is far from as many directions $\bm \eta_j$ as possible in the sense that $\sum_{j=1}^J \exp(-\frac{1}{2} (\bm \eta_j^\top \bm \tau)^2)$ is maximized. This phenomenon is a result of the fact that additive models have significant covariance if $\bm \tau$ is small in any of its directions. For the GAM kernel, the maxima occur on each axis, where $\tilde k(\bm \tau) \ge \frac{d-1}{d}$. That is, covariance does not decay to zero along each axis.

% However, if we draw enough directions $\bm \eta_j$ from an isotropic distribution, a randomly projected additive kernel does decay to zero; $\tilde k(\bm \tau)$ converges to a full-degree, spherical kernel as $J \rightarrow \infty$.
\begin{proposition}\label{prop:convergence}
Let $\phi \colon \R \mapsto [-1, 1]$ be a 1-dimensional kernel, and let $(\bm \eta_j \colon j \ge 1)$ be an i.i.d. sequence of random variables in $\R^d$ drawn from a common isotropic distribution $\mathcal{D}$. Then, for some expected kernel $k_{\text{expected}} \colon \R \mapsto [0,1]$, for any $\bm \tau \in \R^d$, almost surely
\begin{align*}
    \lim_{J\rightarrow \infty} \frac{1}{J} \sum_{j=1}^J \phi(\bm \eta_j^\top \bm \tau)  = \E[\phi(\eta_{1 1} ||\bm \tau||_2)] =: k_{\text{expected}}(||\bm \tau||_2).
\end{align*}

\end{proposition}

For certain choices of sub-kernel $\phi$ and distribution $\mathcal{D}$, we can recover familiar kernels.
\begin{corollary}\label{corollary:conv_to_imq}
If $\phi(x) = e^{-\frac{1}{2}x^2}$ and $\bm \eta_1 \sim \mathcal{N}(0, I_d)$, then
\begin{align} \label{eq:RBF+Gaussian=IMQ}
    k_{\text{expected}}(\bm \tau) = \frac{1}{\sqrt{1+||\bm \tau||_2^2}} \triangleq k_{IMQ}(\bm \tau).
\end{align}
\end{corollary}

\begin{corollary}\label{corollary:conv_to_rbf}
If $\phi(x) = \cos(x)$ and $\bm \eta_1 \sim \mathcal{N}(0, I_d)$, then
\begin{align} \label{eq:cos+Gaussian=RBF}
    k_{\text{expected}}(\bm \tau) = e^{-\frac{1}{2}||\bm \tau||_2^2} \triangleq k_{RBF}(\bm \tau).
\end{align}
\end{corollary}
The expected kernel in (\ref{eq:RBF+Gaussian=IMQ}) is a rational quadratic kernel with parameter $\alpha=1/2$, also known as the \emph{inverse multiquadratic kernel}. It is especially relevant because in this work we focus on the case when the base kernel is RBF. Note that the spectral density provides a standard way to derive sub-kernels associated with higher-dimensional kernels \citep{mantoglou1987digital}.

We can also derive an $O(1/\sqrt{J})$ convergence rate.
\begin{proposition}\label{prop:conv_rate}
Let $\phi$, $k_{\text{expected}}$ be as in Proposition \ref{prop:convergence}. Let $\{\bm \eta_j\}_{j=1}^J$ be a sequence of random variables drawn i.i.d. from an isotropic distribution. Let $\delta > 0$. Then, with probability at least $1-\delta$, we have simultaneously for all pairs of points $\bm \tau_{i,k}$, $i,k\in [n]$, 
\begin{align*}
    \bigg | \frac{1}{J}\sum_{j=1}^J &\phi(\bm \eta_j^\top \bm \tau_{i,k}) - k_{\text{expected}}(||\bm \tau_{i,k}||_2) \bigg | \\ 
    \le  &\frac{2}{3 J} (\log(1/\delta) + 2\log(n) + 1) \\& + \sqrt{\frac{2 \sup_{i,k} \var(\phi(\bm \eta_1^\top \bm \tau_{i,k}))}{J}}
\end{align*}
\end{proposition}
Empirically, we see convergence to the performance of a kernel operating on the original space at a much greater rate. This empirical result is intuitive because even if the resulting kernel after additive random projections is not a multiquadratic kernel, it may still be a good kernel for the data. 

\subsection{Reducing projection redundancy} 

As shown in Proposition \ref{prop:conv_rate}, sampling directions purely randomly converges at the ``slow'' simple Monte Carlo rate of $\mathcal{O}(1/\sqrt{J})$. 
Ideally, we would space directions equally. However, even in only 3-dimensions, this is only possible for certain values of $J=3,15, ...$ \citep{mantoglou1987digital}. In higher dimensions, the problem is highly nontrivial.
%%% Keep for now; the previous text here. %%%
% To minimize redundancy of projections and converge faster, we set $||\bm \eta_j||_2=1$ for each $\bm \eta_j$ and directly maximize a measure of the distance between directions. 
One solution is to numerically maximize a measure of distance between points, such as the antipodal separation distance
\begin{gather*}
    \delta(\bm \eta_1, ..., \bm \eta_J) = \min_{j \ne j^\prime} \cos^{-1}(|\bm \eta_j ^\top \bm \eta_{j^\prime}|),
\end{gather*}
which directly measures the minimal angle between directions.
However, because maximizing $\delta$ is difficult, we instead minimize the loss
\begin{gather} \label{eq:objective}
\ell(\bm \eta_1, ..., \bm \eta_J) = \sum_{j \ne j^\prime} (\bm \eta_j^\top \bm \eta_{j^\prime})^4.
\end{gather}
Minimizing $\ell$ has the effect of increasing the separation distance $\delta$ between directions, though an  optimizer of $\ell$ does not necessarily coincide with an optimizer of $\delta$ unless $d \ge J$.
Additionally, given sufficiently large $J$, a set of directions $\{\bm \eta_j\}_{j=1}^J$ that maximize $\ell$ is a spherical $t$-design with $t=4$ \citep{womersley2018efficient}, thus guaranteeing \textit{optimal order rate decay of worst-case error} for quadrature of smooth functions. 
If $J \le d$, orthogonal directions minimize $\delta$ and $\ell$, so we simply use Gram-Schmidt orthogonalization. Otherwise, we minimize $\ell$ using gradient descent. 
We refer to projected-additive GPs with directions chosen by this method as Diverse Projected-Additive GPs (DPA-GP). We visualize the DPA-GP and other kernels in Figure \ref{fig:covariance-visualization}.

\subsection{Applying length-scales before projection} \label{section:prescale}

If each sub-kernel of an additive kernel learns its own length-scale, it is not clear that the additive kernel approximates an expected kernel. Additionally, if the number of additive kernels $J$ is large, learning separate length-scales introduces many hyperparameters which are also only indirectly related to the original inputs.

Alternatively, we propose applying automatic relevance determination scaling \textit{directly on the original input space} before the data are projected to a low-dimensional space. To learn the length scales, we efficiently propagate gradients through the projections with automatic differentiation. We define
\begin{gather*}
    k_{rp ARD}(\bm x, \bm x^\prime) = \sum_{j=1}^J \alpha_j k_j(P^{(j)} A \bm x, P^{(j)} A \bm x^\prime),
\end{gather*}
where $A = \text{diag}(\bm \sigma^{-1})$. When $\alpha_j = 1/J$ for all $j$ and each $k_j$ has unit length-scales, the theory of section \ref{section:expected_kernel} readily applies, while permitting flexible treatment of length-scales. 

In Section \ref{section:experiments}, we make the empirical discovery that this ARD approach provides significant performance gains.
To distinguish this parameterization from others, we refer to such a model with the \textit{-ARD} suffix.

\subsection{Scaling to large data sets with high dimension with SKI} \label{SKI-Section}

In section \ref{section:background_SKI}, we described how SKI enables scalable GPs, but is constrained to input dimensions of about $d<5$, if no kernel structure is exploited. However, if a kernel decomposes additively by groups of dimensions, it is possible to generalize the applicability of SKI to much higher dimensional spaces. Suppose that a kernel decomposes additively as 
\begin{align*}
    k(\bm x, \bm x^\prime) = \sum_{j=1}^J k_j(\bm x^{(j)}, {\bm x^\prime}^{(j)}),
\end{align*}
where $\bm x^{(j)}$ denotes a group of dimensions of point $\bm x$. Then, the Gram matrix $K$ corresponding to this kernel decomposes similarly, so a matrix-vector multiplication can be performed with each kernel separately as $K \bm v = \sum_{j=1}^J K^{(j)} \bm v$. Since we assume that each sub-kernel $k_j$ is low-degree, each matrix-vector multiplication $K^{(j)} \bm v$ can be computed efficiently using SKI. In particular, in the case that each sub-kernel is 1-dimensional, inference with such a kernel using SKI has complexity $\mathcal{O}(J c (n + m\log m))$, where $m$ is the number of inducing points for each projection and $c$ is the number of iterations of linear conjugate gradients. Typically, $c \ll n$ is sufficient to reach convergence within machine precision, so inference is approximately linear in $n$ \citep{wilson2015kernel}. We demonstrate this asymptotic scaling empirically in Section \ref{section:runtime_experiment}.

\section{Experiments}\label{section:experiments}

\begin{figure*}[h!]
\centering
\begin{subfigure}
\centering\includegraphics[width=5in,height=220pt]{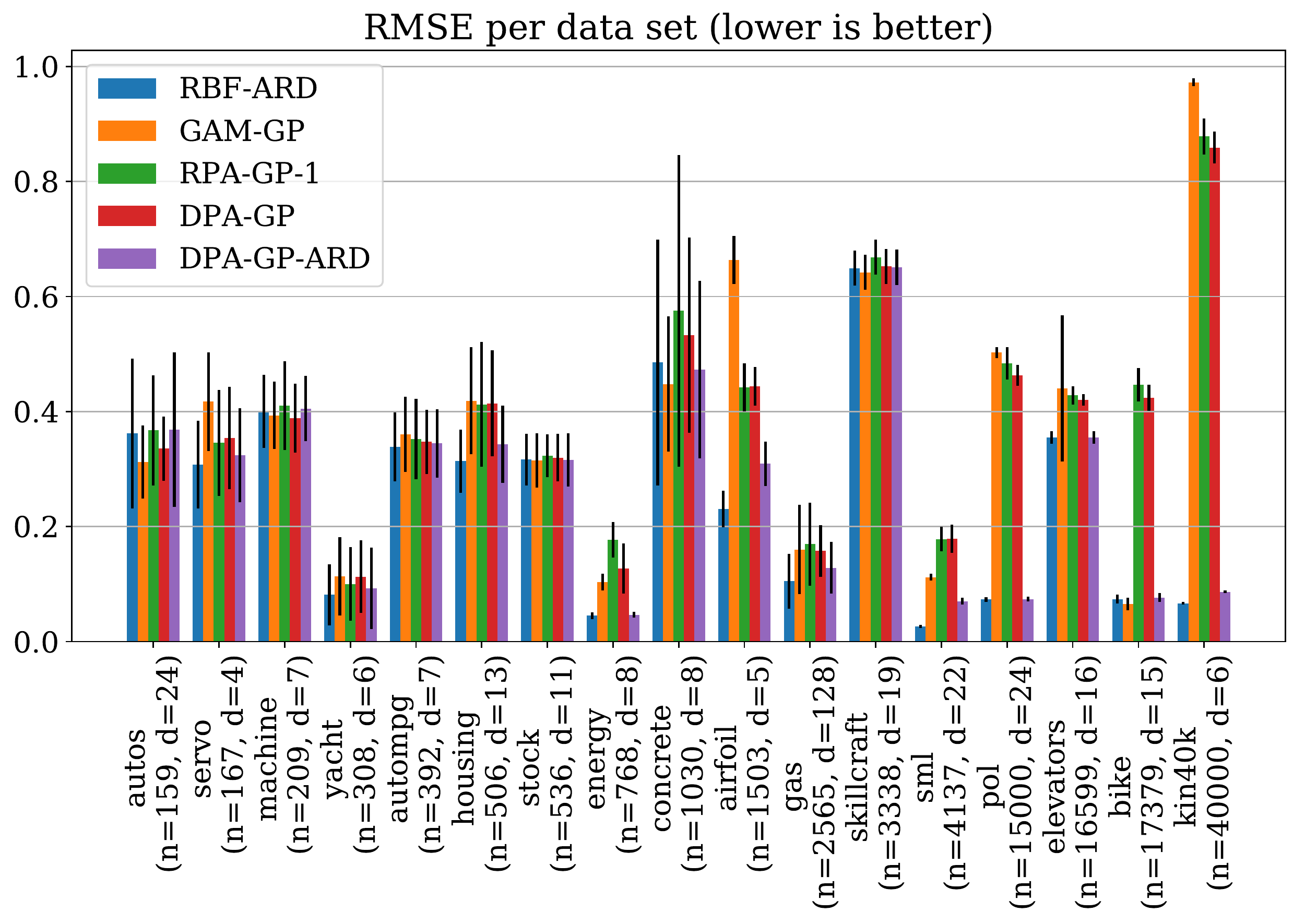}
\caption{The proposed methods and GAM-GP perform surprisingly well compared to RBF-GP. DPA-GP-ARD is able to match the performance of RBF-GP even for large data sets, where the flexibility of GAM-GP begins to be a limiting factor. \label{fig:empirical_gam_vs_dpa}}
\end{subfigure}
\end{figure*}

\renewcommand{\figwidth}{160pt}
\renewcommand{\figheight}{110pt}
\begin{figure*}[h!]
\centering     %%% not \center
\subfigure{\centering \label{fig:concrete}\includegraphics[width=\figwidth, height=\figheight]{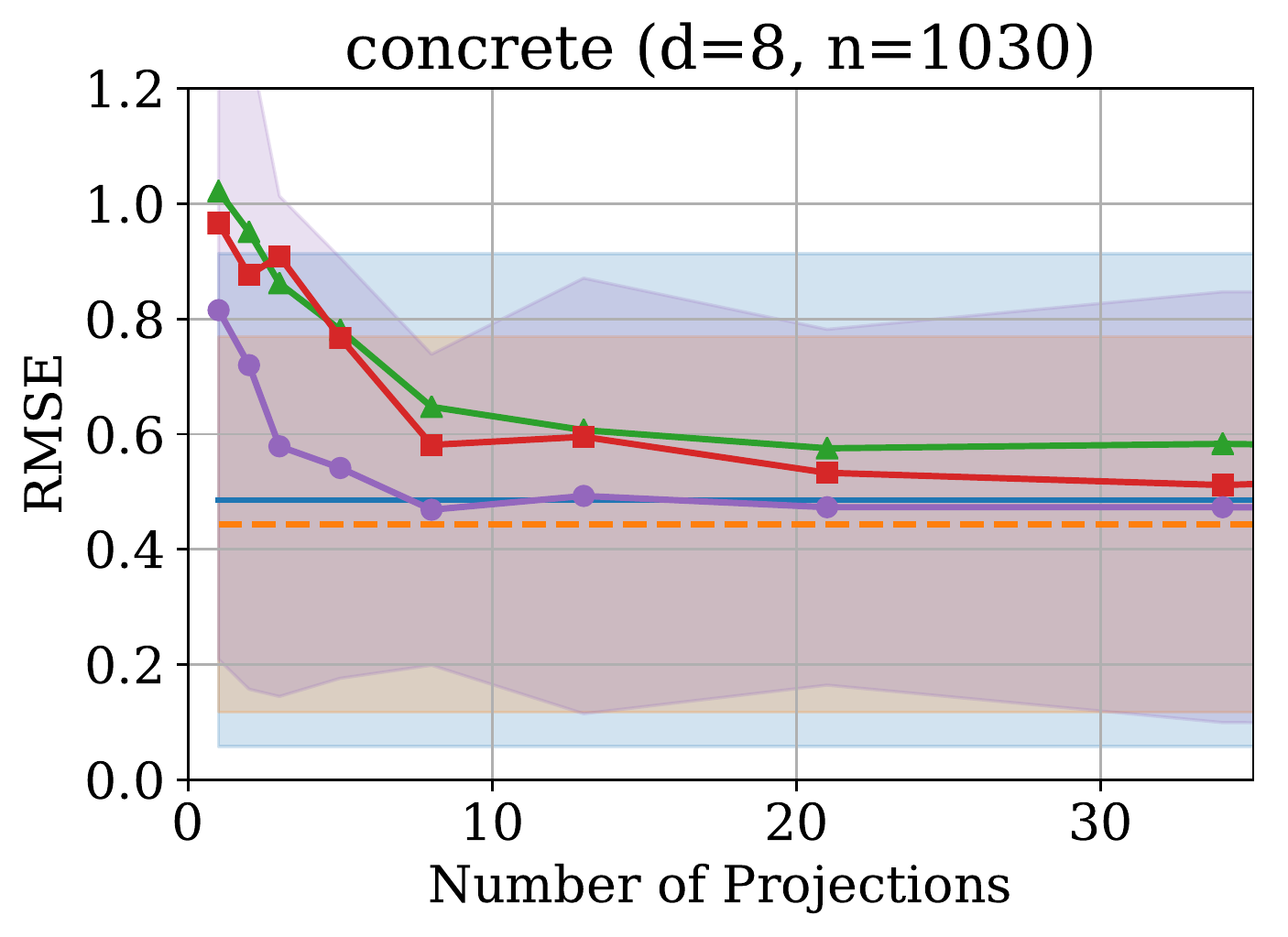}}
% \subfigure{\centering \label{fig:yacht}\includegraphics[width=\figwidth, height=\figheight]{imgs/yacht.pdf}}
\subfigure{\centering \label{fig:autos}\includegraphics[width=\figwidth, height=\figheight]{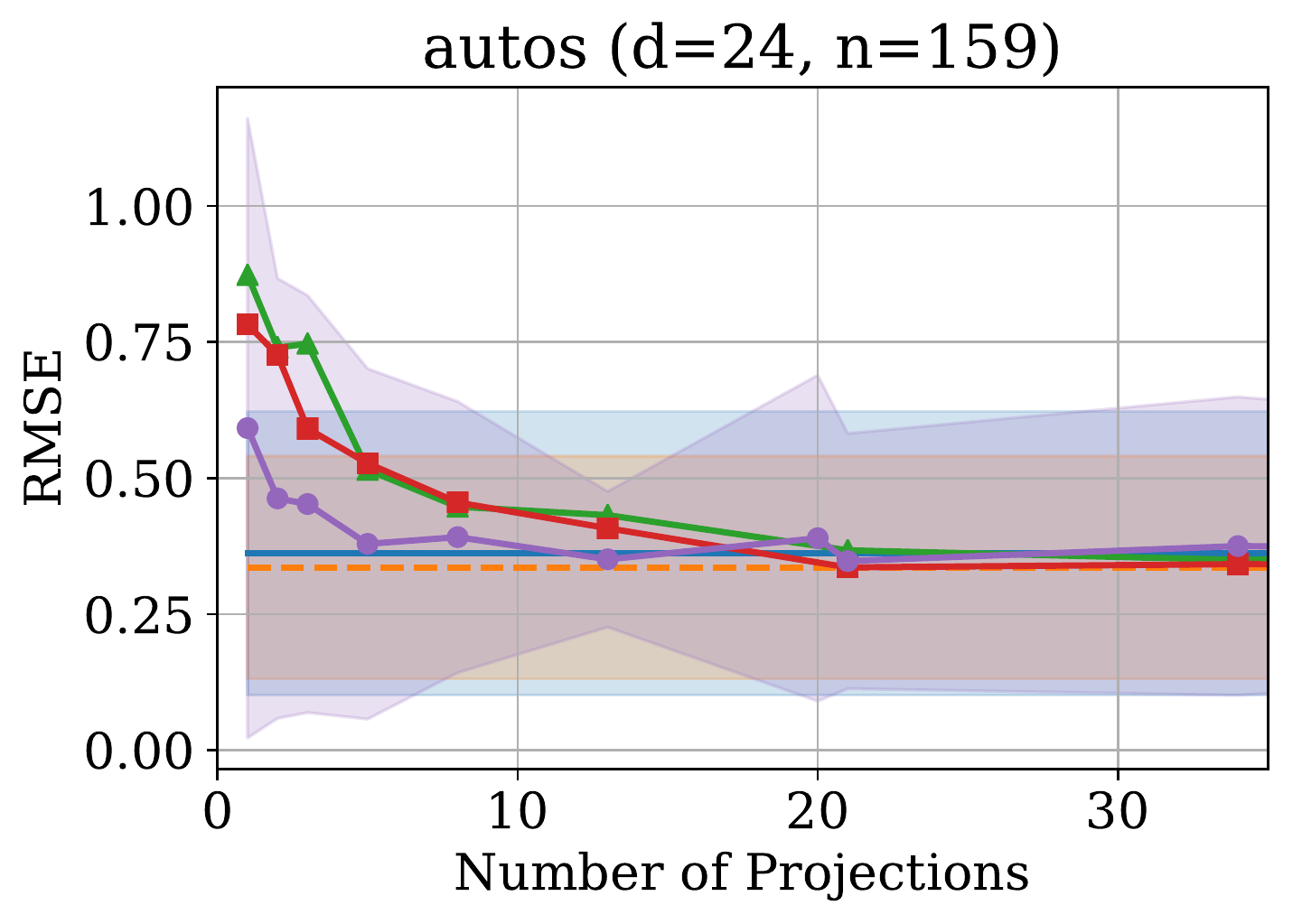}}
\subfigure{\centering \label{fig:housing}\includegraphics[width=\figwidth, height=\figheight]{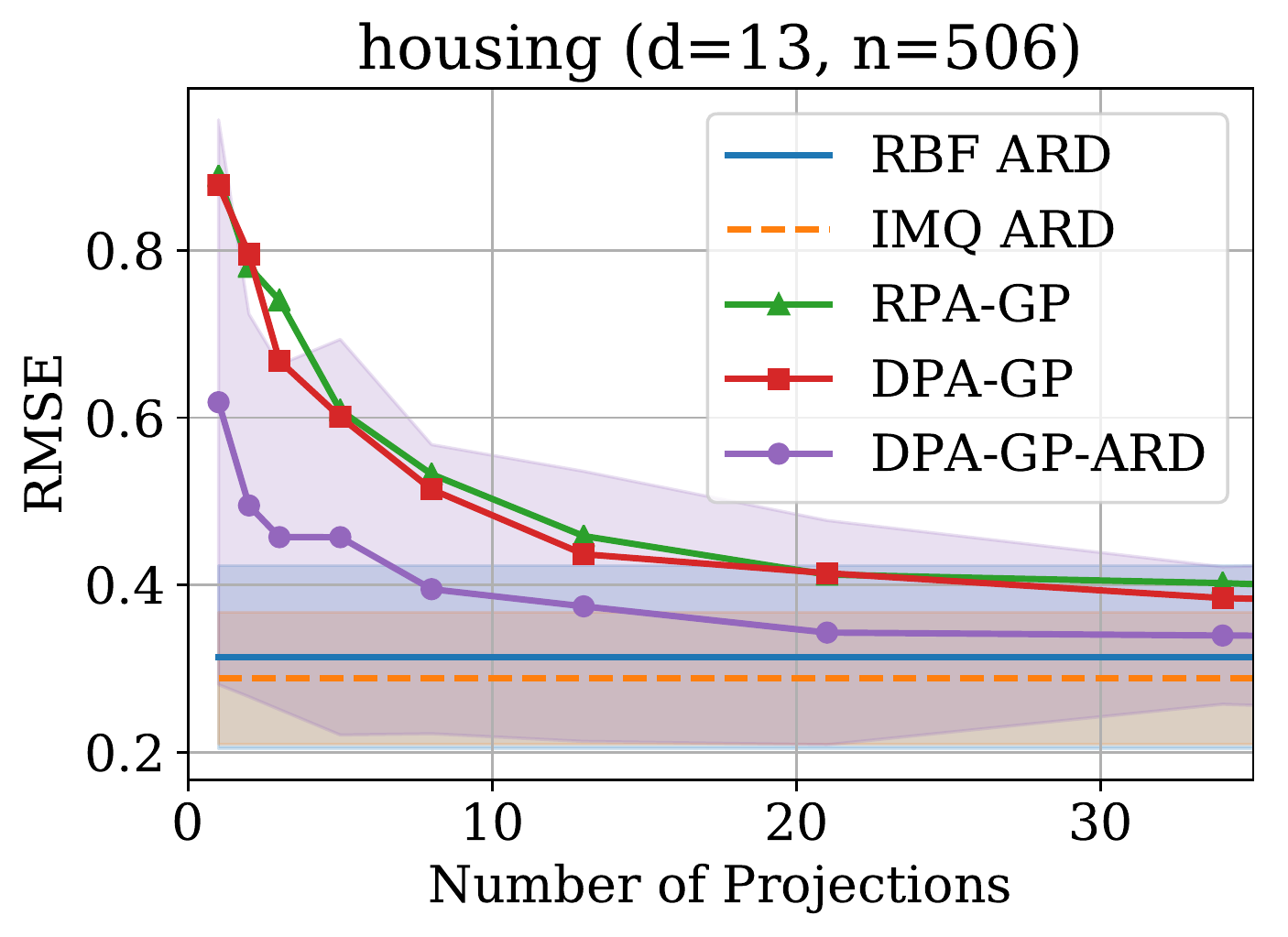}}

\caption{Representative test RMSE of RPA-GP and DPA-GP as the number of projections vary compared to full-dimensional RBF and inverse multiquadratic (IMQ) kernels. Shaded regions are 2 times the standard deviation over cross-validation, and lines are the average RMSE. For clarity, we only show the variation for DPA-GP-ARD. In general, there is a fast convergence to the performance of RBF and IMQ kernels, and DPA-GP consistently improves upon RPA-GP by a small amount, and applying length-scales before (DPA-GP-ARD) projection dramatically increases performance.\label{Convergence}}
\end{figure*}

We evaluate RPA-GP and DPA-GP on a wide array of regression tasks. We compare the predictive accuracy of the proposed methods to GPs with RBF and GAM kernels (section \ref{section:regression_benchmarks}), study the effect of increasing number of projections (section \ref{section:ablation_on_J}), compare predictive accuracy under various assumptions via synthetic data sets and on very high-dimensional data (section \ref{section:gam_comparisons}), and demonstrate the superior asymptotic scaling of RPA-GP with SKI over traditional inference (section \ref{section:runtime_experiment}). We implement all models using GPyTorch \citep{gardner2018gpytorch} and provide code at \url{https://github.com/idelbrid/Randomly-Projected-Additive-GPs}.

\subsection{Benchmarks on UCI data sets} \label{section:regression_benchmarks}
To evaluate RPA-GP, we compare each method's regression performance on a large number of UCI data sets. For each model, and for each data set, we perform 10-fold cross validation twice to accurately measure the performance of stochastic methods. For each fold, we normalize the features and target function by mean and standard deviation as computed on the training folds, so predicting the mean results in $\approx 1$ RMSE. For each fold, we fit the kernel hyperparameters by maximizing the log-marginal likelihood. We use Adam \citep{kingma2014adam} with learning rate 0.1 for at most 1000 iterations, stopping if log marginal likelihood improves by less than 0.0001 over 20 iterations, smoothing with a moving average. We use a smoothed box prior over the likelihood noise parameter to encourage numerical stability. 
We compute the root mean squared error on the held-out fold for each model, data set, and fold, and report the RMSE in Figure \ref{fig:empirical_gam_vs_dpa}. 
To reiterate, \texttt{RBF-ARD} is a standard RBF GP with ARD; \texttt{GAM-GP} is an additive GP with RBF subkernels and ARD. \texttt{RPA-GP-1}, \texttt{DPA-GP}, and \texttt{DPA-GP-ARD} are additive across 20 1-dimensional projections. \texttt{RPA-GP-1} uses Gaussian random projections; \texttt{DPA-GP} minimizes the objective in Equation \ref{eq:objective}; and \texttt{DPA-GP-ARD} performs the pre-projection scaling method described in Section \ref{section:prescale}.

We report results for additional models tested in the appendix. In particular, using SKI for scalable inference, enabled by additive projections, results in essentially identical performance as exact inference. Additionally, an IMQ kernel achieves error similar to an RBF kernel. Further, we find as expected that including sub-kernels of slightly higher degree can improve performance.

\subsection{Convergence of random projected-additive GP accuracy} \label{section:ablation_on_J}

To study the sensitivity of RPA-GP to the number of projections and the behavior as $J\rightarrow \infty$, we perform an ablation study for several data sets. For each data set, we perform 10-fold cross validation (twice) with 1-degree random projected-additive GPs and vary the number of projections. We show representative plots in Figure \ref{Convergence}. In these experiments, the benefit of DPA-GP-ARD also becomes obvious, as we see it is able to converge much more quickly than the other approaches. By the time $J=20$, represented in Figure \ref{fig:empirical_gam_vs_dpa}, the various projection approaches are more comparable.

\subsection{Comparisons to fully-additive kernel}
\label{section:gam_comparisons}

\begin{figure}[h]
    \centering
    \includegraphics[width=180pt,height=\figheight]{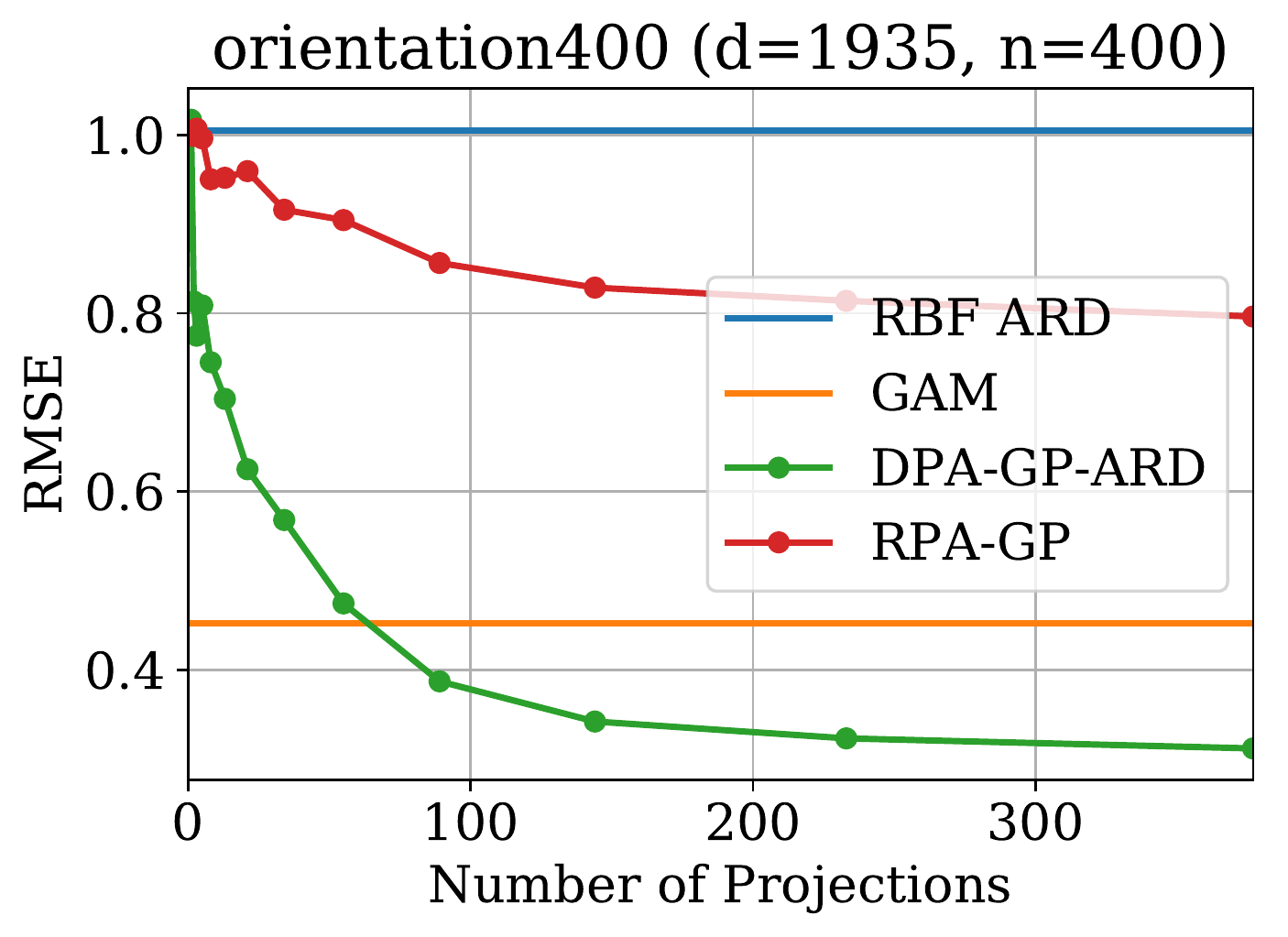}
    \caption{Comparative performance on the Olivetti face orientation regression task. With high dimensionality and a non-additive target function, DPA-GP-ARD outperforms alternatives, though the number of projections $J$ must be somewhat high.}
    \label{fig:just_orientation400}
\end{figure}
\begin{figure}[h]
    \centering
    \includegraphics[width=180pt,height=\figheight]{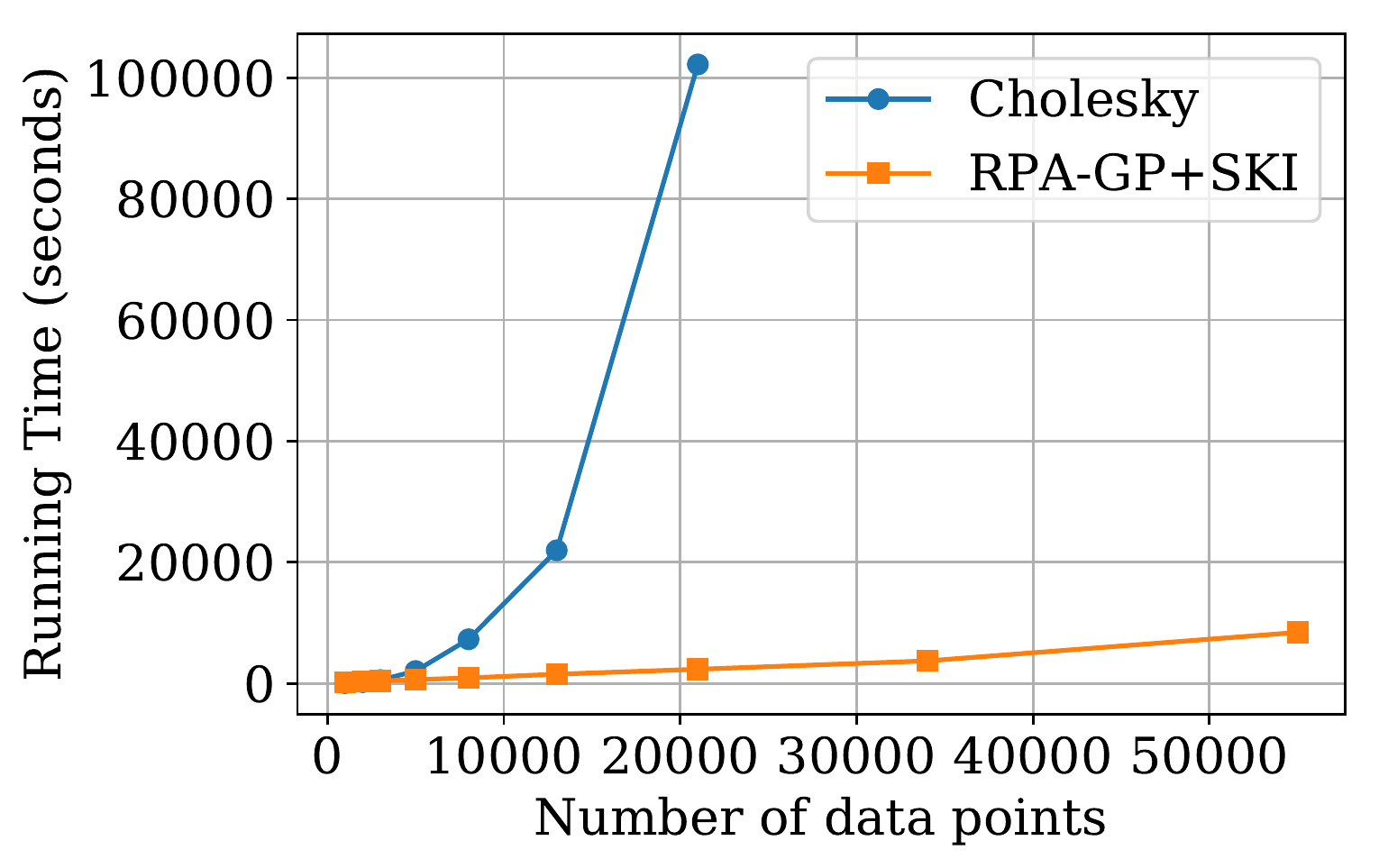}
    \caption{Training run time of RPA-GP with SKI compared to canonical Cholesky decomposition-based inference on synthetic data with 100 input dimensions and varying numbers of points following $x \sim N(0,I_d),$ $y=\sum_{i=1}^{100} \sin(x_i)+\epsilon$.  We note that runtime behaviour is somewhat data dependent, due to changes in conditioning of the kernel matrix, but the scaling of SKI remains approximately linear, compared to the cubic scaling of the Cholesky decomposition.
    \label{fig:ski_scaling_figure}}
\end{figure}

\renewcommand{\figwidth}{196pt}
\renewcommand{\figheight}{120pt}
\begin{figure*}[tb]
\centering
\begin{subfigure}
\centering\includegraphics[width=\figwidth,height=\figheight]{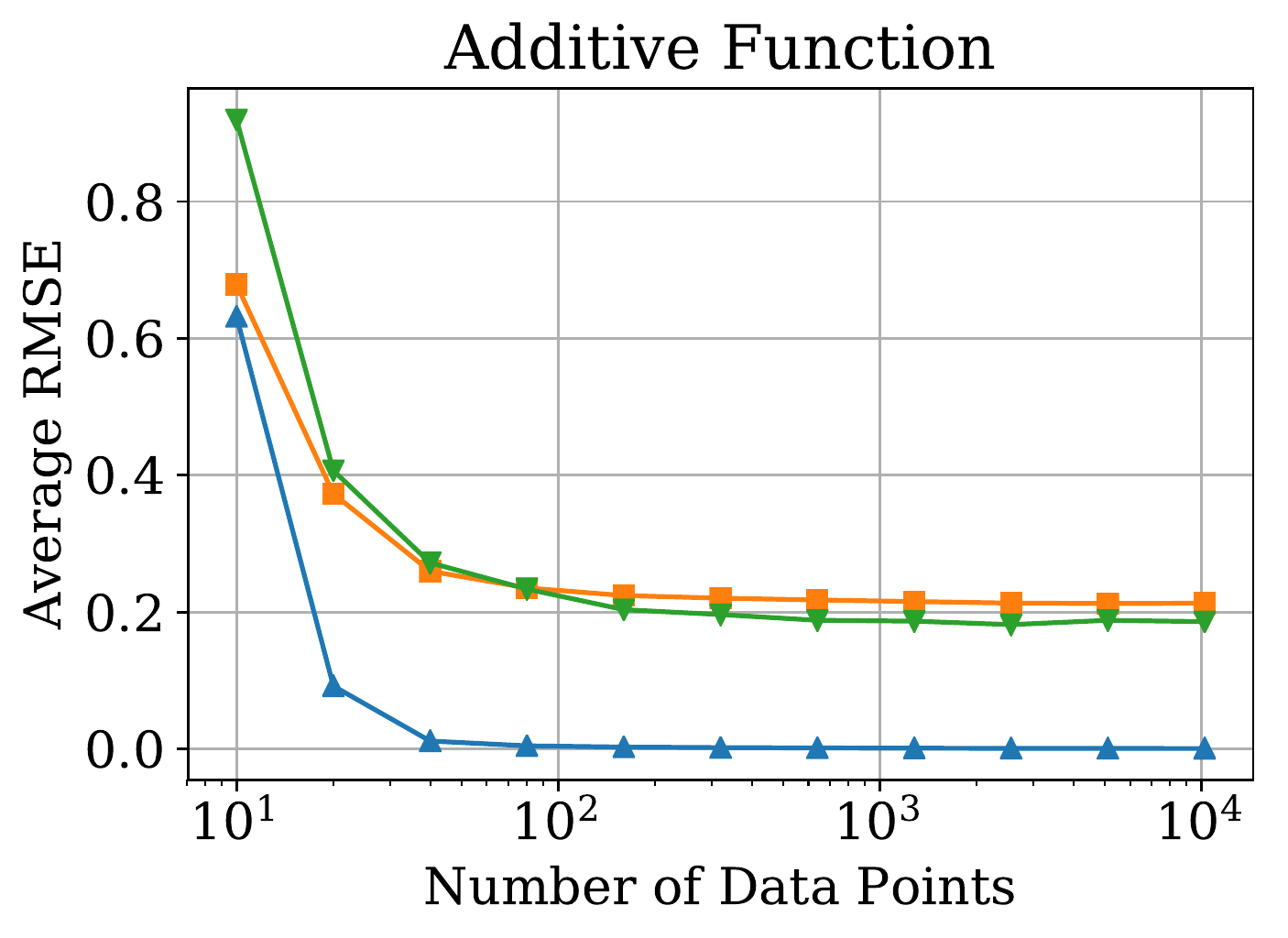}
\end{subfigure}
\begin{subfigure}
\centering\includegraphics[width=\figwidth,height=\figheight]{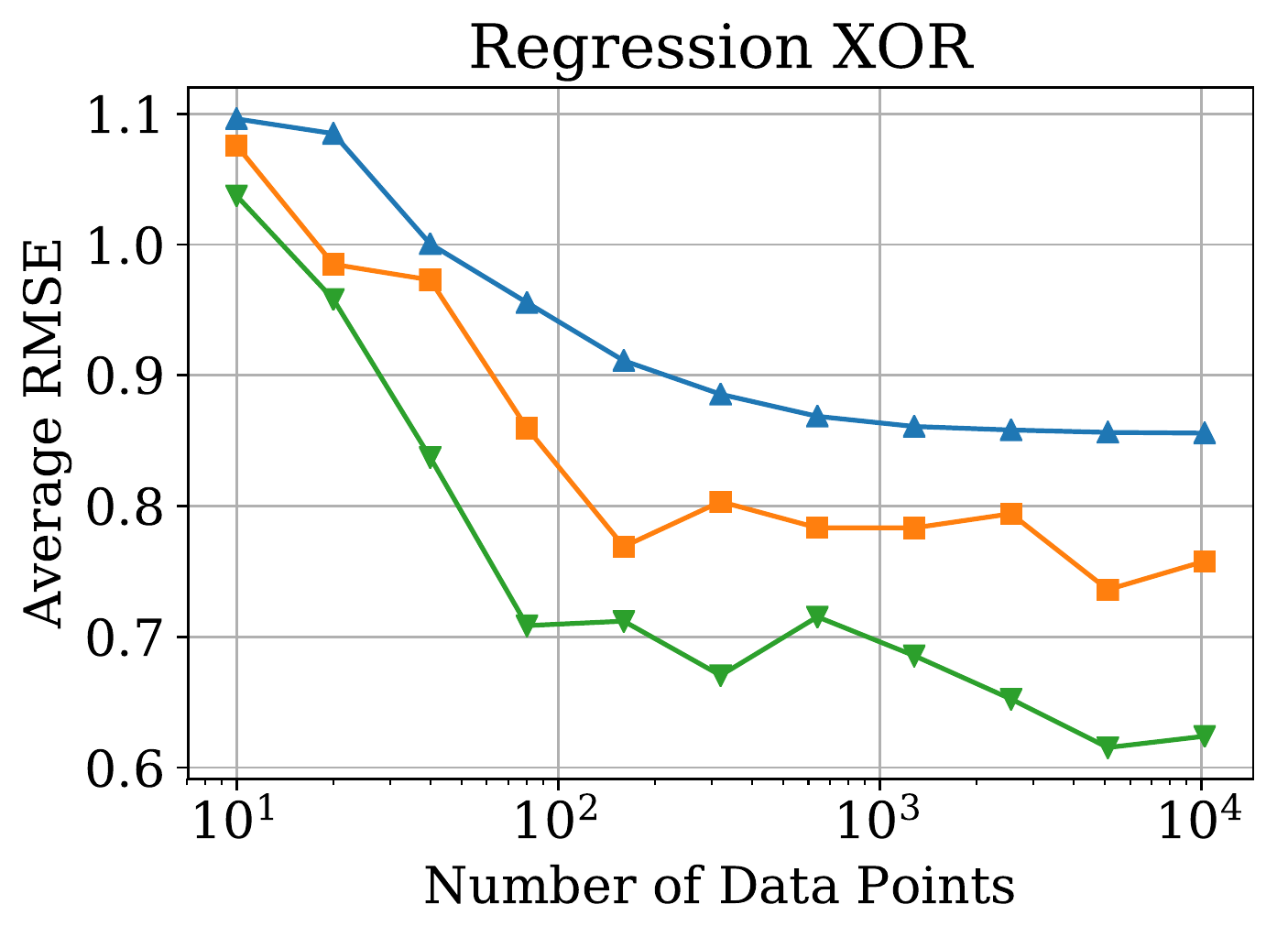}
\end{subfigure}
\begin{subfigure}
\centering\includegraphics[width=\figwidth,height=\figheight]{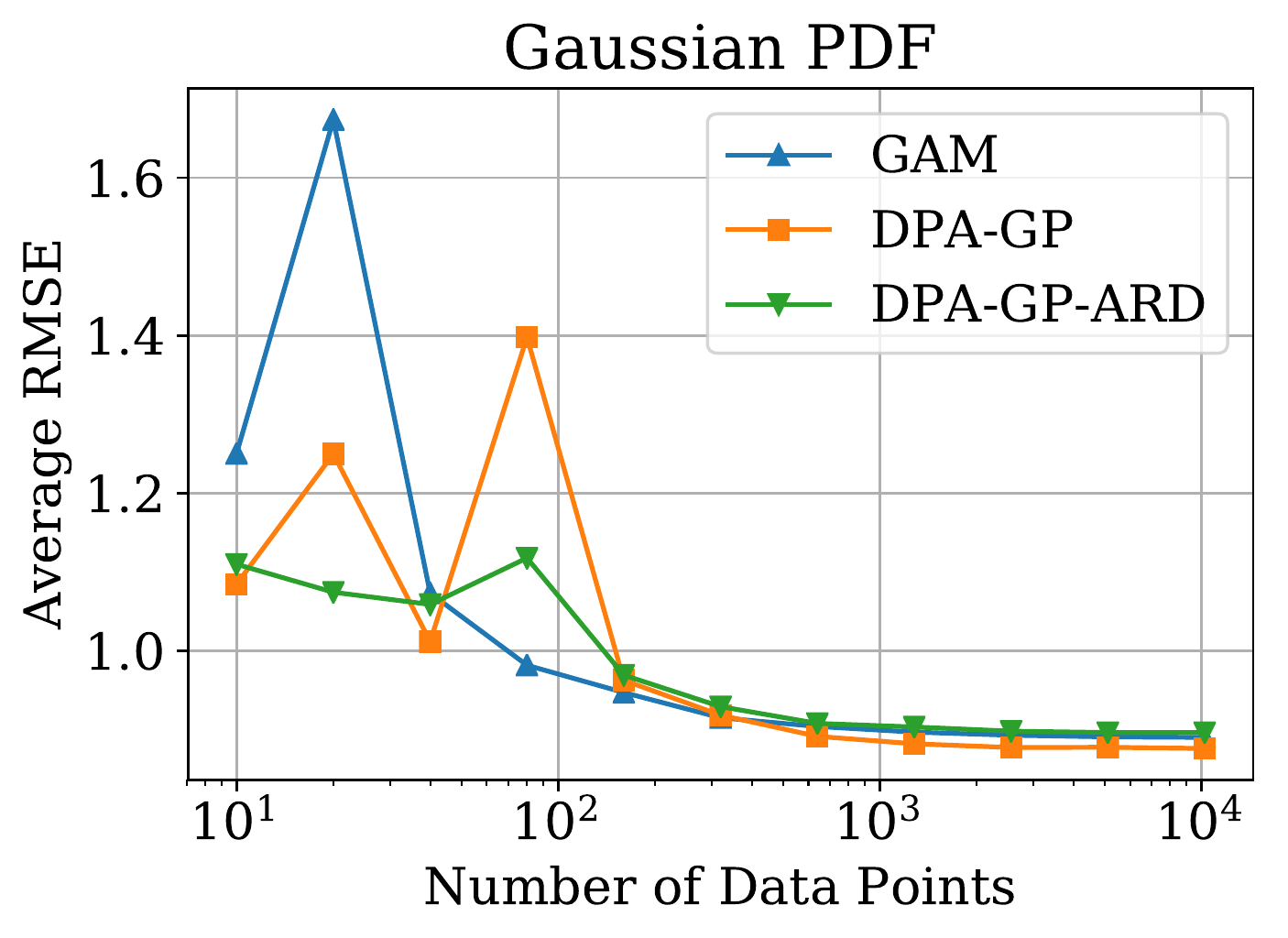}
\end{subfigure}
\begin{subfigure}
\centering\includegraphics[width=\figwidth,height=\figheight]{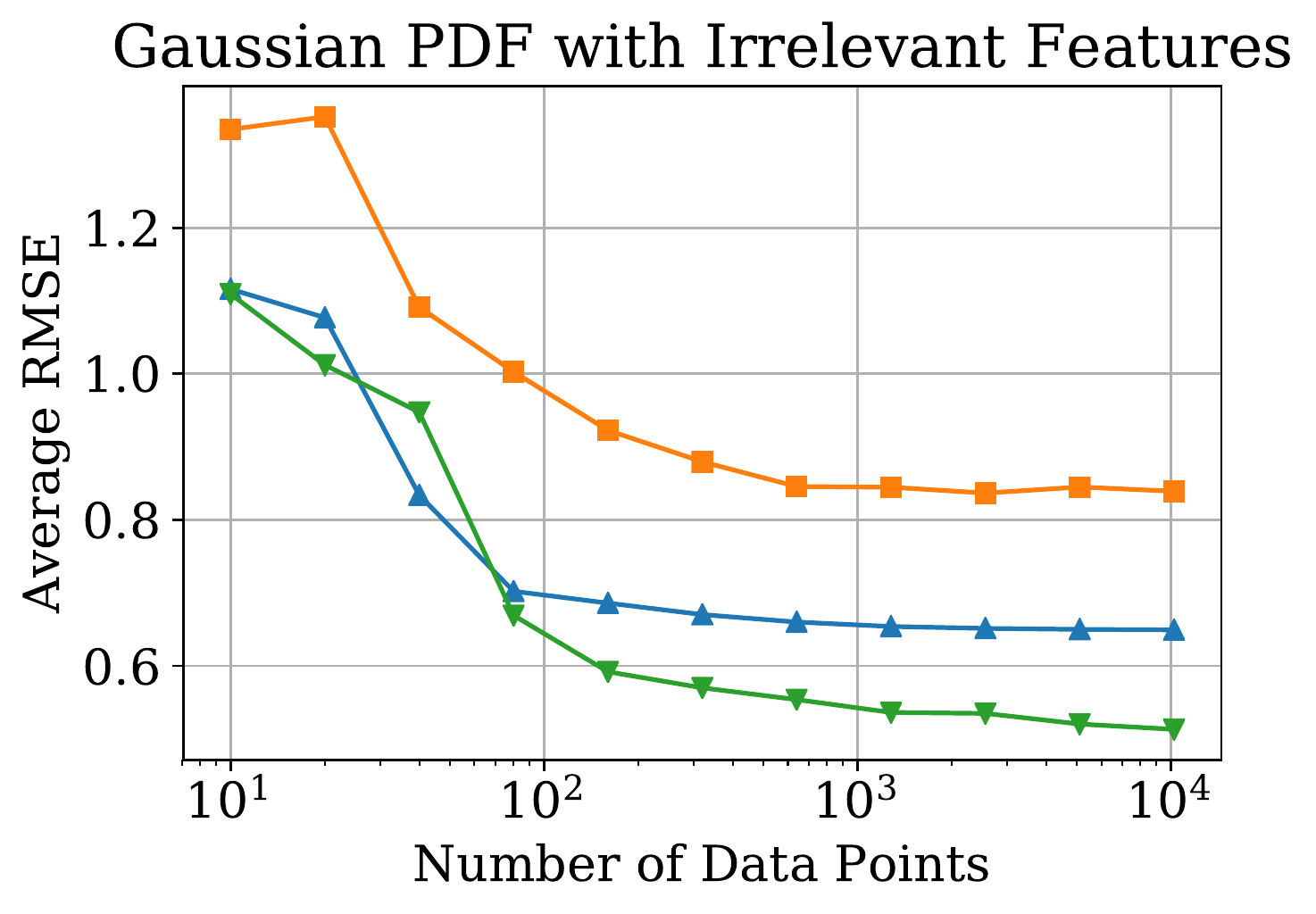}
\end{subfigure}
\caption{\textbf{Top left}: Average RMSE on an additive function. GAM is best as expected. \textbf{Top right}: Average RMSE on a very non-additive function: a smooth relaxation of the XOR function. GAM performs worse than DPA methods. \textbf{Bottom left}: Average RMSE synthetic data with a rotation-invariant function. All methods perform equally. \textbf{Bottom right}: Average RMSE synthetic data with half irrelevant features. Plain DPA-GP cannot distinguish relevant features from irrelevant ones after projection, but DPA-GP-ARD improves on both DPA-GP and GAM.
\label{fig:synthetic_gam_vs_dpa}}
\end{figure*}

As observed in section \ref{section:regression_benchmarks}, GAM GP performs surprisingly well in comparison to the standard RBF-ARD kernel, despite its limited model class. This result is noteworthy in its own right,
since the GAM GP ignores everything but first order interactions between inputs. To our knowledge it is not known that such a parsimonious representation can achieve comparable results to a kernel acting on the full inputs over this significant range of experiments.

The GAM kernel is also  related to DPA-GP. Using $J=d$ 1-dimensional projections, DPA-GP is equivalent to a GAM with a random rotation applied to input data (though the ability to tune model expressiveness by controlling $J$ is a key benefit to a random projection approach). As a result, we expect the methods to perform similarly if the target function does not vary mostly in axis-aligned directions.

To understand the differences between GAM GP and our proposed techniques, we consider additional empirical tests on synthetic data and very high-dimensional data sets.

\textbf{Synthetic regression tasks}: 
In Figure \ref{fig:synthetic_gam_vs_dpa}, we test each method as we increase the number of data points on additive synthetic data. We see that GAM, as expected, performs better for target functions that are truly additive. Conversely, GAM can perform poorly if the function is not additive. For target functions that are rotation-invariant, all of GAM, DPA-GP, and DPA-GP-ARD perform equivalently as expected. However, if irrelevant features are introduced, which we conjecture is a frequent occurrence in real regression problems, DPA-GP-ARD performs better than GAM. Irrelevant features introduce noise into the projections, but ARD prunes irrelevant features and effectually reduces the input dimension, wherein DPA-GP-ARD still uses $d$ additive components, but GAM then effectually uses $<d$ additive components. Hence, DPA-GP-ARD provides a better approximation of a non-additive kernel.

\textbf{High dimensional regression tasks}: 
We construct regression data sets of three different sizes from the Olivetti faces data set, following \citet{wilson2016deep} and \citet{NIPS2007_3211}. We uniformly subsample images, uniformly sample a rotation in $[-90, 90]$, crop the rotated images, and use the rotations as regression targets. DPA-GP-ARD outperforms RBF and GAM GPs with when $n$ is small compared to $d$. Results with $n=400$ images are presented in Figure \ref{fig:just_orientation400}. We additionally test on three genomics data sets, finding that DPA-GP-ARD and GAM GP generally perform comparably and provide figures in the appendix.

\subsection{Scaling to large data sets} \label{section:runtime_experiment}

To demonstrate the asymptotic computational complexity of RPA-GP with SKI, we train both RPA-GP with SKI and a GP with RBF kernel using Cholesky-based inference for 120 Adam iterations on data sets constructed by drawing
$X \sim \mathcal{N}(0, I_d)$, $\bm y_t = \sum_{i=1}^{d} \sin(x_i^t) + \epsilon_t$, $\epsilon_t \sim 0.01 \cdot \mathcal{N}(0, 1)$ 
with $d=100$ and a varying number data points. We use RPA-GP with 20 1-dimensional projections and 512 inducing points per projection. We run this experiment on a MacBook Air with a 1.8 GHz Intel i5 processor with 8 GB of RAM. The results are presented in Figure \ref{fig:ski_scaling_figure}.

Note that it is infeasible to run SKI without RPA-GP with a reasonable number of inducing points on data sets with this number of input dimensions; even if $d=6$ and we have $100$ inducing points in each dimension, the resulting $1$ trillion inducing points represented as 32-bit floating point numbers cannot be stored in memory. 

\section{Conclusion}\label{discussion}

We proposed novel learning-free algorithms to construct additive Gaussian processes by using sums of low-dimensional kernels operating over random (RPA-GP) projections. We demonstrated the remarkable result that these approaches achieve the performance of kernels operating over the full-dimensional input space even when projecting into a \emph{one-dimensional} space and \emph{without learning} the projections. 

Moreover, we proved that by taking enough additive random projections, RPA-GP converges to the inverse multiquadratic kernel and proposed a novel deterministic algorithm (DPA-GP) to reduce projection redundancy that indeed improves regression performance. We demonstrated that a fully-additive GP also achieves remarkable empirical performance and illustrated that, while such a model is superior if its assumption holds, combining DPA-GP with automatic relevance determination directly on input features (DPA-GP-ARD) performs as well as fully additive GPs on very high-dimensional data and better on large data sets.  Finally, as an added benefit, we demonstrated that by exploiting the additive structure of RPA-GP, we essentially reduce inference from a $d$ dimensional problem to $J$ 1-dimensional problems, enabling the application of SKI \citep{wilson2015kernel} and thereby reducing standard GP computational complexity from $\mathcal{O}(n^3)$ to $\mathcal{O}(J c (n + m))$, where $J$ is the number of random projections, $c$ is the number of linear conjugate gradients iterations, and $m$ is the number of inducing points. These results are of great practical significance: we have shown that GP regression, which is the backbone of many procedures, can often be effectively reduced to one-dimension without requiring the training of a projection. These approaches also naturally generalize the applicability of popular scalable inference procedures, such as SKI, which have been conventionally constrained to lower dimensional spaces.

In short, we demonstrate the pleasing result that a range of regression problems can be reduced to a \emph{single} input dimension, while retaining or even improving accuracy, without having to learn a projection. In a single dimension, methods become much easier to analyze and scale, leading a rich variety of future research directions.

\subsubsection*{Acknowledgements}
AGW and IAD were supported by Amazon Research Award, Facebook Research,
NSF IIS-1563887, and NSF IIS-1910266. We thank Alex Smola for helpful discussions.

\bibliographystyle{plainnat}
\bibliography{references}

\begin{thebibliography}{45}
\providecommand{\natexlab}[1]{#1}
\providecommand{\url}[1]{\texttt{#1}}
\expandafter\ifx\csname urlstyle\endcsname\relax
  \providecommand{\doi}[1]{doi: #1}\else
  \providecommand{\doi}{doi: \begingroup \urlstyle{rm}\Url}\fi

\bibitem[Ahmed(2004)]{ahmed2004multiple}
Yousuf~Shamim Ahmed.
\newblock \emph{Multiple random projection for fast, approximate nearest
  neighbor search in high dimensions}.
\newblock University of Toronto, 2004.

\bibitem[Bach(2009)]{bach2009high}
Francis Bach.
\newblock High-dimensional non-linear variable selection through hierarchical
  kernel learning.
\newblock \emph{arXiv preprint arXiv:0909.0844}, 2009.

\bibitem[Baraniuk and Wakin(2009)]{baraniuk2009random}
Richard~G Baraniuk and Michael~B Wakin.
\newblock Random projections of smooth manifolds.
\newblock \emph{Foundations of computational mathematics}, 9\penalty0
  (1):\penalty0 51--77, 2009.

\bibitem[Bull(2011)]{bull2011convergence}
Adam~D Bull.
\newblock Convergence rates of efficient global optimization algorithms.
\newblock \emph{Journal of Machine Learning Research}, 12\penalty0
  (Oct):\penalty0 2879--2904, 2011.

\bibitem[Cheney and Light(2009)]{cheney2009course}
Elliott~Ward Cheney and William~Allan Light.
\newblock \emph{A course in approximation theory}, volume 101.
\newblock American Mathematical Soc., 2009.

\bibitem[Christakos(1987)]{christakos1987stochastic}
George Christakos.
\newblock Stochastic simulation of spatially correlated geo-processes.
\newblock \emph{Mathematical Geology}, 19\penalty0 (8):\penalty0 807--831,
  1987.

\bibitem[Deisenroth and Rasmussen(2011)]{deisenroth2011pilco}
Marc Deisenroth and Carl~E Rasmussen.
\newblock Pilco: A model-based and data-efficient approach to policy search.
\newblock In \emph{Proceedings of the 28th International Conference on machine
  learning (ICML-11)}, pages 465--472, 2011.

\bibitem[Dong et~al.(2017)Dong, Eriksson, Nickisch, Bindel, and
  Wilson]{dong2017scalable}
Kun Dong, David Eriksson, Hannes Nickisch, David Bindel, and Andrew~G Wilson.
\newblock Scalable log determinants for gaussian process kernel learning.
\newblock In \emph{Advances in Neural Information Processing Systems}, pages
  6330--6340, 2017.

\bibitem[Duvenaud(2014)]{duvenaud2014automatic}
David Duvenaud.
\newblock \emph{Automatic model construction with Gaussian processes}.
\newblock PhD thesis, University of Cambridge, 2014.

\bibitem[Duvenaud et~al.(2013)Duvenaud, Lloyd, Grosse, Tenenbaum, and
  Zoubin]{duvenaud2013structure}
David Duvenaud, James Lloyd, Roger Grosse, Joshua Tenenbaum, and Ghahramani
  Zoubin.
\newblock Structure discovery in nonparametric regression through compositional
  kernel search.
\newblock In Sanjoy Dasgupta and David McAllester, editors, \emph{Proceedings
  of the 30th International Conference on Machine Learning}, volume~28 of
  \emph{Proceedings of Machine Learning Research}, pages 1166--1174, Atlanta,
  Georgia, USA, 17--19 Jun 2013. PMLR.
\newblock URL \url{http://proceedings.mlr.press/v28/duvenaud13.html}.

\bibitem[Duvenaud et~al.(2011)Duvenaud, Nickisch, and
  Rasmussen]{duvenaud2011additive}
David~K Duvenaud, Hannes Nickisch, and Carl~E Rasmussen.
\newblock Additive gaussian processes.
\newblock In \emph{Advances in neural information processing systems}, pages
  226--234, 2011.

\bibitem[Engel et~al.(2005)Engel, Mannor, and Meir]{Engel2005reinforcement}
Yaakov Engel, Shie Mannor, and Ron Meir.
\newblock Reinforcement learning with gaussian processes.
\newblock In \emph{Proceedings of the 22Nd International Conference on Machine
  Learning}, ICML '05, pages 201--208, New York, NY, USA, 2005. ACM.
\newblock ISBN 1-59593-180-5.
\newblock \doi{10.1145/1102351.1102377}.
\newblock URL \url{http://doi.acm.org/10.1145/1102351.1102377}.

\bibitem[Freulon and Lantuejoul(1993)]{freulon1993revisiting}
Xavier Freulon and Christian Lantuejoul.
\newblock Revisiting the turning bands method.
\newblock \emph{Acta Stereologica}, 1993.

\bibitem[Friedman and Stuetzle(1981)]{friedman1981projection}
Jerome~H Friedman and Werner Stuetzle.
\newblock Projection pursuit regression.
\newblock \emph{Journal of the American statistical Association}, 76\penalty0
  (376):\penalty0 817--823, 1981.

\bibitem[Gardner et~al.(2017)Gardner, Guo, Weinberger, Garnett, and
  Grosse]{gardner2017discovering}
Jacob Gardner, Chuan Guo, Kilian Weinberger, Roman Garnett, and Roger Grosse.
\newblock Discovering and exploiting additive structure for bayesian
  optimization.
\newblock In \emph{Artificial Intelligence and Statistics}, pages 1311--1319,
  2017.

\bibitem[Gardner et~al.(2018)Gardner, Pleiss, Weinberger, Bindel, and
  Wilson]{gardner2018gpytorch}
Jacob Gardner, Geoff Pleiss, Kilian~Q Weinberger, David Bindel, and Andrew~G
  Wilson.
\newblock Gpytorch: Blackbox matrix-matrix gaussian process inference with gpu
  acceleration.
\newblock In \emph{Advances in Neural Information Processing Systems}, pages
  7576--7586, 2018.

\bibitem[Gilboa et~al.(2013)Gilboa, Saat{\c{c}}i, and
  Cunningham]{gilboa2013scaling}
Elad Gilboa, Yunus Saat{\c{c}}i, and John Cunningham.
\newblock Scaling multidimensional gaussian processes using projected additive
  approximations.
\newblock In \emph{International Conference on Machine Learning}, pages
  454--461, 2013.

\bibitem[Gneiting(1998)]{gneiting1998closed}
Tilmann Gneiting.
\newblock Closed form solutions of the two-dimensional turning bands equation.
\newblock \emph{Mathematical Geology}, 30\penalty0 (4):\penalty0 379--390,
  1998.

\bibitem[Guhaniyogi and Dunson(2016)]{guhaniyogi2016compressed}
Rajarshi Guhaniyogi and David~B Dunson.
\newblock Compressed gaussian process for manifold regression.
\newblock \emph{The Journal of Machine Learning Research}, 17\penalty0
  (1):\penalty0 2472--2497, 2016.

\bibitem[Hastie and Tibshirani(1986)]{hastie1986}
Trevor Hastie and Robert Tibshirani.
\newblock Generalized additive models.
\newblock \emph{Statist. Sci.}, 1\penalty0 (3):\penalty0 297--310, 08 1986.
\newblock \doi{10.1214/ss/1177013604}.
\newblock URL \url{https://doi.org/10.1214/ss/1177013604}.

\bibitem[Herlands et~al.(2019)Herlands, Neill, Nickisch, and
  Wilson]{herlands2018change}
William Herlands, Daniel~B. Neill, Hannes Nickisch, and Andrew~Gordon Wilson.
\newblock Change surfaces for expressive multidimensional changepoints and
  counterfactual prediction.
\newblock \emph{Journal of Machine Learning Research}, 20\penalty0
  (99):\penalty0 1--51, 2019.
\newblock URL \url{http://jmlr.org/papers/v20/17-352.html}.

\bibitem[Hinton and Salakhutdinov(2008)]{NIPS2007_3211}
Geoffrey~E Hinton and Ruslan~R Salakhutdinov.
\newblock Using deep belief nets to learn covariance kernels for gaussian
  processes.
\newblock In J.~C. Platt, D.~Koller, Y.~Singer, and S.~T. Roweis, editors,
  \emph{Advances in Neural Information Processing Systems 20}, pages
  1249--1256. Curran Associates, Inc., 2008.

\bibitem[Kandasamy et~al.(2015)Kandasamy, Schneider, and
  P{\'o}czos]{kandasamy2015high}
Kirthevasan Kandasamy, Jeff Schneider, and Barnab{\'a}s P{\'o}czos.
\newblock High dimensional bayesian optimisation and bandits via additive
  models.
\newblock In \emph{International Conference on Machine Learning}, pages
  295--304, 2015.

\bibitem[Keys(1981)]{keys1981cubic}
Robert Keys.
\newblock Cubic convolution interpolation for digital image processing.
\newblock \emph{IEEE transactions on acoustics, speech, and signal processing},
  29\penalty0 (6):\penalty0 1153--1160, 1981.

\bibitem[Kingma and Ba(2014)]{kingma2014adam}
Diederik~P Kingma and Jimmy Ba.
\newblock Adam: A method for stochastic optimization.
\newblock \emph{arXiv preprint arXiv:1412.6980}, 2014.

\bibitem[Lantu{\'e}joul(2013)]{lantuejoul2013geostatistical}
Christian Lantu{\'e}joul.
\newblock \emph{Geostatistical simulation: models and algorithms}.
\newblock Springer Science \& Business Media, 2013.

\bibitem[Li et~al.(2016)Li, Kandasamy, Poczos, and Schneider]{li2016high}
Chun-Liang Li, Kirthevasan Kandasamy, Barnabas Poczos, and Jeff Schneider.
\newblock High dimensional bayesian optimization via restricted projection
  pursuit models.
\newblock In Arthur Gretton and Christian~C. Robert, editors, \emph{Proceedings
  of the 19th International Conference on Artificial Intelligence and
  Statistics}, volume~51 of \emph{Proceedings of Machine Learning Research},
  pages 884--892, Cadiz, Spain, 09--11 May 2016. PMLR.
\newblock URL \url{http://proceedings.mlr.press/v51/li16e.html}.

\bibitem[Mantoglou(1987)]{mantoglou1987digital}
Aristotelis Mantoglou.
\newblock Digital simulation of multivariate two-and three-dimensional
  stochastic processes with a spectral turning bands method.
\newblock \emph{Mathematical Geology}, 19\penalty0 (2):\penalty0 129--149,
  1987.

\bibitem[Mantoglou and Wilson(1982)]{mantoglou1982turning}
Aristotelis Mantoglou and John~L Wilson.
\newblock The turning bands method for simulation of random fields using line
  generation by a spectral method.
\newblock \emph{Water Resources Research}, 18\penalty0 (5):\penalty0
  1379--1394, 1982.

\bibitem[Matheron(1973)]{matheron197.intrinsic}
G.~Matheron.
\newblock The intrinsic random functions and their applications.
\newblock \emph{Advances in Applied Probability}, 5\penalty0 (3):\penalty0
  439--468, 1973.
\newblock ISSN 00018678.
\newblock URL \url{http://www.jstor.org/stable/1425829}.

\bibitem[Mo{\v{c}}kus(1975)]{movckus1975bayesian}
Jonas Mo{\v{c}}kus.
\newblock On bayesian methods for seeking the extremum.
\newblock In \emph{Optimization Techniques IFIP Technical Conference}, pages
  400--404. Springer, 1975.

\bibitem[Qamar and Tokdar(2014)]{qamar2014additive}
Shaan Qamar and Surya~T Tokdar.
\newblock Additive gaussian process regression.
\newblock \emph{arXiv preprint arXiv:1411.7009}, 2014.

\bibitem[Rasmussen and Williams(2006)]{rasmussen2006gaussian}
Carl~Edward Rasmussen and Christopher~KI Williams.
\newblock \emph{Gaussian processes for machine learning}, volume~2.
\newblock MIT Press Cambridge, MA, 2006.

\bibitem[Saat{\c{c}}i(2012)]{saatcci2012scalable}
Yunus Saat{\c{c}}i.
\newblock \emph{Scalable inference for structured Gaussian process models}.
\newblock PhD thesis, Citeseer, 2012.

\bibitem[Sarlos(2006)]{sarlos2006improved}
Tamas Sarlos.
\newblock Improved approximation algorithms for large matrices via random
  projections.
\newblock In \emph{2006 47th Annual IEEE Symposium on Foundations of Computer
  Science (FOCS'06)}, pages 143--152. IEEE, 2006.

\bibitem[Srinivas et~al.(2010)Srinivas, Krause, Kakade, and
  Seeger]{Srinivas2019Gaussian}
Niranjan Srinivas, Andreas Krause, Sham Kakade, and Matthias Seeger.
\newblock Gaussian process optimization in the bandit setting: No regret and
  experimental design.
\newblock In \emph{Proceedings of the 27th International Conference on
  International Conference on Machine Learning}, ICML'10, pages 1015--1022,
  USA, 2010. Omnipress.
\newblock ISBN 978-1-60558-907-7.
\newblock URL \url{http://dl.acm.org/citation.cfm?id=3104322.3104451}.

\bibitem[Stone et~al.(1985)]{stone1985additive}
Charles~J Stone et~al.
\newblock Additive regression and other nonparametric models.
\newblock \emph{The annals of Statistics}, 13\penalty0 (2):\penalty0 689--705,
  1985.

\bibitem[{van Beers} and {Kleijnen}(2004)]{vanBeers2004kriging}
W.~C.~M. {van Beers} and J.~P.~C. {Kleijnen}.
\newblock Kriging interpolation in simulation: a survey.
\newblock In \emph{Proceedings of the 2004 Winter Simulation Conference,
  2004.}, volume~1, page 121, Dec 2004.
\newblock \doi{10.1109/WSC.2004.1371308}.

\bibitem[Wang et~al.(2017)Wang, Gehring, Kohli, and Jegelka]{wang2017batched}
Zi~Wang, Clement Gehring, Pushmeet Kohli, and Stefanie Jegelka.
\newblock Batched large-scale bayesian optimization in high-dimensional spaces.
\newblock In \emph{AISTATS}, 2017.

\bibitem[Wang et~al.(2016)Wang, Hutter, Zoghi, Matheson, and
  de~Feitas]{wang2016bayesian}
Ziyu Wang, Frank Hutter, Masrour Zoghi, David Matheson, and Nando de~Feitas.
\newblock Bayesian optimization in a billion dimensions via random embeddings.
\newblock \emph{Journal of Artificial Intelligence Research}, 55:\penalty0
  361--387, 2016.

\bibitem[Williams and Rasmussen(1996)]{williams1996gaussian}
Christopher~KI Williams and Carl~Edward Rasmussen.
\newblock Gaussian processes for regression.
\newblock In \emph{Advances in neural information processing systems}, pages
  514--520, 1996.

\bibitem[Wilson and Adams(2013)]{wilson2013gaussian}
Andrew Wilson and Ryan Adams.
\newblock Gaussian process kernels for pattern discovery and extrapolation.
\newblock In \emph{International Conference on Machine Learning}, pages
  1067--1075, 2013.

\bibitem[Wilson and Nickisch(2015)]{wilson2015kernel}
Andrew Wilson and Hannes Nickisch.
\newblock Kernel interpolation for scalable structured gaussian processes
  (kiss-gp).
\newblock In \emph{International Conference on Machine Learning}, pages
  1775--1784, 2015.

\bibitem[Wilson et~al.(2016)Wilson, Hu, Salakhutdinov, and
  Xing]{wilson2016deep}
Andrew~Gordon Wilson, Zhiting Hu, Ruslan Salakhutdinov, and Eric~P Xing.
\newblock Deep kernel learning.
\newblock In \emph{Artificial Intelligence and Statistics}, pages 370--378,
  2016.

\bibitem[Womersley(2018)]{womersley2018efficient}
Robert~S Womersley.
\newblock Efficient spherical designs with good geometric properties.
\newblock In \emph{Contemporary Computational Mathematics-A Celebration of the
  80th Birthday of Ian Sloan}, pages 1243--1285. Springer, 2018.

\end{thebibliography}
\clearpage
\appendix 
\appendixpage
\section{Proofs}

\subsection{Proof of Proposition \ref{prop:convergence}} \label{appendix}
\begin{proof}
Because $\bm \eta_1$ is drawn from an isotropic distribution, we can let $\bm \tau = [x, 0, 0, ..., 0]$ without loss of generality. Then, define the expected kernel value at $\bm \tau$ as the expectation
\begin{align*}
    \E[k(\bm \eta_1^\top \bm \tau)] = \E[k(\eta_{1 1} ||\bm \tau||_2)] =: k_{\text{expected}}(||\bm \tau||_2)
\end{align*}
Then, by the Law of Large Numbers, the empirical mean $\frac{1}{J} \sum_{j=1}^J k(\bm \eta_j^\top \bm \tau)$ converges to the expectation $k_{\text{expected}}(||\bm \tau||_2)$ almost surely as $J \rightarrow \infty$. 
\end{proof}

\subsection{Proof of Corollary \ref{corollary:conv_to_imq}}
\begin{proof}
Let $r = ||\bm \tau||_2$ for simplicity.

Suppose $\bm \eta_1 \sim \mathcal{N}(0, I_d)$m which implies $\eta_{1 1} \sim \mathcal{N}(0, 1)$. Then,  
\begin{align*}
    \E[e^{-\frac{1}{2}\eta_{1 1}^2r^2}] &= \int_{-\infty}^\infty \frac{1}{\sqrt{2 \pi}} e^{-\frac{1}{2}\eta_{1 1}^2r^2} e^{-\frac{1}{2} \eta_{1 1}^2} d\eta_{ 1 1} \\
    &= \int_{-\infty}^\infty \frac{1}{\sqrt{2 \pi}} e^{-\frac{1}{2} \eta_{1 1}^2 (1 + r^2)} \\
    &= \frac{1}{\sqrt{1 + r^2}}.
\end{align*}
\end{proof}

\subsection{Proof of Corollary \ref{corollary:conv_to_rbf}}
\begin{proof}
Let $r = ||\bm \tau||_2$.

Suppose $\bm \eta_1 \sim \mathcal{N}(0, I_d)$ which implies $\eta_{1 1} \sim \mathcal{N}(0, 1)$. Then,  
\begin{align*}
    \E [\cos(\eta_{1 1} r)] &= 
    \E \left[\sum_{j=0}^\infty \frac{(-1)^{j}}{(2j)!} (\eta_{1 1} r)^{2j} \right]  \\
    &= \sum_{j=0}^\infty \frac{(-1)^j}{(2j)!} r^{2j} \E[\eta_{1 1}^{2 j}] \\
    &= \sum_{j=0}^\infty \frac{(-1)^j}{(2j)!} r^{2j} (2j-1)!!  \\
    &= \sum_{j=0}^\infty \frac{(-1)^j}{(2j)!!} r^{2j} \\
    &= \sum_{j=0}^\infty \frac{(-1)^j}{2^{j} j!} (r^2)^{j} \\
    &= e^{- \frac{1}{2} r^{2}}.
\end{align*}
\end{proof} 

\subsection{Proof of Proposition 2}

\begin{proof}
From Proposition 1, we have an empirical mean of 1-dimensional kernels approximating its mean. We can apply concentration inequalities. We choose Bernstein's inequality to explain the effect of the projection variance on convergence:

\begin{align} \label{bernstein}
    &P\left(\bigg| \frac{1}{J} \sum_{j=1}^J \phi(\bm \eta_j^\top \bm \tau) - k_{exp}(\bm \tau) \bigg| \ge \epsilon \right) \\
    \le &\exp\left( \frac{-\epsilon^2 J}{2v(\bm \tau) + 4 \epsilon /3} \right) 
\end{align}

Letting the right-hand size equal $\delta > 0$ and solving for $\epsilon$, we have

\begin{align*}
    \delta &= \exp\left( \frac{-\epsilon^2 J}{2v(\bm \tau) + 4 \epsilon /3} \right) \\
    \log(1/\delta) &= \frac{\epsilon^2 J}{2 v(\bm \tau) + 4 \epsilon / 3} \\
    \left(\frac{1}{J}\log(1/\delta) \right)(2 v(\bm \tau) + 4 \epsilon / 3) &= \epsilon^2 \\
    \frac{2v(\bm \tau)}{J}\log(1/\delta) + \frac{4 }{3 J}\log(1/\delta) \epsilon &= \epsilon^2
    \\
    \epsilon^2 + b \epsilon + c &= 0 
\end{align*}
for $b = -4/(3J)\log(1/\delta)$, $c = -2 v(\bm \tau)/J \log(1/\delta)$. Then, 
\begin{align*}
    \epsilon &= \frac{-b \pm \sqrt{b^2 -4 c }}{2a} \\
    &= \frac{4/(3J)\log(1/\delta) \pm \sqrt{16/(9J^2) + 8 v(\bm \tau) /J}}{2} \\
    &= 2/(3J)\log(1/\delta) + \sqrt{4/(9J^2) + 2 v(\bm \tau) /J} \\
    &\le \frac{2}{3 J} (\log(1/\delta)+1) + \sqrt{\frac{2 v(\bm \tau)}{J}}
\end{align*}

Finally, to derive the uniform convergence bound, applying union bounds to equation \ref{bernstein} and following similar simplification steps leads to the bound

\begin{align*}
\epsilon \le \frac{2}{3 J} (\log(1/\delta) + 2\log(n) + 1) + \sqrt{\frac{2 \sup_{i,k} v(\bm \tau_{i,k})}{J}}
\end{align*}
\end{proof}

\section{Supplementary Results}

We show the regression performance of additional models from Section \ref{section:regression_benchmarks} in Table \ref{table:supplemental test table}. 

\begin{table*}[]

\fontsize{8pt}{0.8em}\selectfont
\centering
\begin{tabular}{@{}l|ll|ccccc@{}}
\toprule
dataset & n & d & RBF-ARD & IMQ-ARD & Single-RP & RPA-GP-2 & RPA-GP-3\\ \hline 
challenger & 23 & 4 & $\bm{1.04 \pm 1.28$} & $\bm{1.00 \pm 1.33$} & $\bm{0.97 \pm 1.20$} & $\bm{1.16 \pm 1.48$} & $\bm{1.24 \pm 1.64$}\\ 
fertility & 100 & 9 & $\bm{1.02 \pm 0.43$} & $\bm{0.94 \pm 0.37$} & $\bm{0.99 \pm 0.49$} & $\bm{0.98 \pm 0.40$} & $\bm{1.05 \pm 0.46$}\\ 
concreteslump & 103 & 7 & $\bm{0.11 \pm 0.13$} & $\bm{0.10 \pm 0.11$} & $0.99 \pm 0.59$ & $\bm{0.11 \pm 0.12$} & $\bm{0.09 \pm 0.09$}\\ 
autos & 159 & 24 & $\bm{0.36 \pm 0.26$} & $\bm{0.34 \pm 0.20$} & $0.83 \pm 0.28$ & $\bm{0.33 \pm 0.26$} & $\bm{0.35 \pm 0.14$}\\ 
servo & 167 & 4 & $\bm{0.31 \pm 0.15$} & $\bm{0.31 \pm 0.16$} & $0.90 \pm 0.21$ & $\bm{0.32 \pm 0.16$} & $\bm{0.35 \pm 0.18$}\\ 
breastcancer & 194 & 33 & $\bm{0.98 \pm 0.34$} & $\bm{0.90 \pm 0.27$} & $0.99 \pm 0.18$ & $\bm{0.94 \pm 0.31$} & $0.98 \pm 0.29$\\ 
machine & 209 & 7 & $\bm{0.40 \pm 0.13$} & $\bm{0.39 \pm 0.09$} & $0.82 \pm 0.47$ & $\bm{0.39 \pm 0.13$} & $\bm{0.38 \pm 0.15$}\\ 
yacht & 308 & 6 & $\bm{0.08 \pm 0.11$} & $0.08 \pm 0.11$ & $0.87 \pm 0.35$ & $\bm{0.07 \pm 0.11$} & $\bm{0.07 \pm 0.10$}\\ 
autompg & 392 & 7 & $\bm{0.34 \pm 0.12$} & $\bm{0.34 \pm 0.13$} & $0.71 \pm 0.37$ & $\bm{0.35 \pm 0.12$} & $\bm{0.34 \pm 0.15$}\\ 
housing & 506 & 13 & $\bm{0.31 \pm 0.11$} & $\bm{0.29 \pm 0.08$} & $0.93 \pm 0.38$ & $0.37 \pm 0.15$ & $0.36 \pm 0.17$\\ 
forest & 517 & 12 & $\bm{1.06 \pm 0.40$} & $\bm{1.02 \pm 0.39$} & $\bm{0.99 \pm 0.39$} & $\bm{1.06 \pm 0.37$} & $\bm{1.07 \pm 0.37$}\\ 
stock & 536 & 11 & $\bm{0.32 \pm 0.09$} & $\bm{0.32 \pm 0.07$} & $0.84 \pm 0.33$ & $\bm{0.32 \pm 0.08$} & $\bm{0.33 \pm 0.10$}\\ 
energy & 768 & 8 & $\bm{0.05 \pm 0.01$} & $0.05 \pm 0.01$ & $0.84 \pm 0.33$ & $0.07 \pm 0.05$ & $\bm{0.04 \pm 0.02$}\\ 
concrete & 1030 & 8 & $\bm{0.49 \pm 0.43$} & $\bm{0.44 \pm 0.33$} & $0.94 \pm 0.66$ & $\bm{0.58 \pm 0.75$} & $0.55 \pm 0.45$\\ 
airfoil & 1503 & 5 & $0.23 \pm 0.06$ & $\bm{0.20 \pm 0.04$} & $0.92 \pm 0.17$ & $0.24 \pm 0.05$ & $\bm{0.20 \pm 0.03$}\\ 
gas & 2565 & 128 & $\bm{0.11 \pm 0.10$} & $\bm{0.10 \pm 0.11$} & $0.68 \pm 0.41$ & $0.13 \pm 0.09$ & $\bm{0.13 \pm 0.11$}\\ 

\end{tabular}

\vspace{.5cm}
\centering
\begin{tabular}{@{}l|ll|ccccc@{}}
dataset & n & d & RPA-GP-1 & DPA-GP & RPA-GP-ARD & DPA-GP-ARD & DPA-GP-ARD-SKI\\ \hline 
challenger & 23 & 4 & $\bm{0.93 \pm 1.45$} & $\bm{1.02 \pm 1.47$} & $\bm{1.02 \pm 1.28$} & $\bm{0.98 \pm 1.30$} & $\bm{0.98 \pm 1.30$}\\ 
fertility & 100 & 9 & $1.09 \pm 0.51$ & $\bm{1.02 \pm 0.45$} & $\bm{1.05 \pm 0.42$} & $\bm{0.95 \pm 0.42$} & $\bm{0.99 \pm 0.42$}\\ 
concreteslump & 103 & 7 & $\bm{0.10 \pm 0.09$} & $0.09 \pm 0.08$ & $0.14 \pm 0.26$ & $\bm{0.10 \pm 0.08$} & $\bm{0.10 \pm 0.08$}\\ 
autos & 159 & 24 & $\bm{0.37 \pm 0.19$} & $\bm{0.34 \pm 0.11$} & $\bm{0.36 \pm 0.19$} & $\bm{0.37 \pm 0.27$} & $\bm{0.36 \pm 0.22$}\\ 
servo & 167 & 4 & $\bm{0.35 \pm 0.18$} & $0.35 \pm 0.18$ & $\bm{0.34 \pm 0.19$} & $\bm{0.32 \pm 0.16$} & $\bm{0.34 \pm 0.16$}\\ 
breastcancer & 194 & 33 & $1.03 \pm 0.27$ & $\bm{0.90 \pm 0.31$} & $1.04 \pm 0.26$ & $1.13 \pm 0.26$ & $1.00 \pm 0.30$\\ 
machine & 209 & 7 & $\bm{0.41 \pm 0.15$} & $\bm{0.39 \pm 0.12$} & $\bm{0.40 \pm 0.10$} & $\bm{0.41 \pm 0.11$} & $\bm{0.40 \pm 0.11$}\\ 
yacht & 308 & 6 & $\bm{0.10 \pm 0.13$} & $0.11 \pm 0.13$ & $\bm{0.09 \pm 0.12$} & $\bm{0.09 \pm 0.14$} & $0.10 \pm 0.13$\\ 
autompg & 392 & 7 & $\bm{0.35 \pm 0.14$} & $\bm{0.35 \pm 0.11$} & $\bm{0.36 \pm 0.14$} & $\bm{0.34 \pm 0.12$} & $\bm{0.34 \pm 0.11$}\\ 
housing & 506 & 13 & $0.41 \pm 0.22$ & $0.41 \pm 0.18$ & $0.38 \pm 0.13$ & $0.34 \pm 0.13$ & $0.38 \pm 0.17$\\ 
forest & 517 & 12 & $\bm{1.03 \pm 0.36$} & $\bm{1.01 \pm 0.37$} & $\bm{1.06 \pm 0.39$} & $\bm{1.05 \pm 0.35$} & $\bm{1.01 \pm 0.37$}\\ 
stock & 536 & 11 & $\bm{0.32 \pm 0.07$} & $\bm{0.32 \pm 0.08$} & $\bm{0.32 \pm 0.08$} & $\bm{0.32 \pm 0.09$} & $\bm{0.32 \pm 0.08$}\\ 
energy & 768 & 8 & $0.18 \pm 0.06$ & $0.13 \pm 0.09$ & $0.05 \pm 0.01$ & $\bm{0.05 \pm 0.01$} & $0.06 \pm 0.02$\\ 
concrete & 1030 & 8 & $0.58 \pm 0.54$ & $0.53 \pm 0.34$ & $\bm{0.46 \pm 0.36$} & $\bm{0.47 \pm 0.31$} & $\bm{0.49 \pm 0.31$}\\ 
airfoil & 1503 & 5 & $0.44 \pm 0.08$ & $0.44 \pm 0.07$ & $0.32 \pm 0.06$ & $0.31 \pm 0.08$ & $0.32 \pm 0.08$\\ 
gas & 2565 & 128 & $0.17 \pm 0.14$ & $0.16 \pm 0.09$ & $0.13 \pm 0.08$ & $0.13 \pm 0.09$ & - \\ 
\bottomrule
\end{tabular}

\caption{\label{table:supplemental test table}Average RMSE on UCI regression data sets with 2 times standard deviation. For each data set, we bold the best model and models whose means are not statistically significantly different according to a 1-sided $t$-test against the best model. Models are described in Table \ref{table:supplemental model explanation}. RBF-ARD and IMQ-ARD are generally the best models and perform similarly. A single random projection is handily beaten by all additive random projection methods. Among 1-dimensional random projections, there are slight benefits to using a diverse projected additive (DPA) GP. There is a large benefit to applying length-scales before projection (-ARD). There is little to no performance loss using SKI. Finally, from RPA-GP-2 and RPA-GP-3, there is a benefit for adding more random projections and sub-kernels of higher-degrees. The last experiment, DPA-GP-ARD-SKI on gas was not completed in time, but we fully believe it will continue the pattern.}  
\end{table*}

\begin{table*} \label{table:supplemental model explanation}
\centering
\fontsize{8pt}{0.8em}\selectfont
\begin{tabular}{l|cccc}
model  & sub-kernel & projection method & sub-kernel degrees & pre-scale?\\ \hline
RBF-ARD & RBF & - & $1 \times d$ & - \\
IMQ-ARD & Inverse Multiquadratic & - & $1 \times d$ & - \\
Single-RP & RBF & Gaussian & $1 \times 1$ & No \\
RPA-GP-2 & RBF & Gaussian & $4 \times 1, 4 \times 2, 4 \times 3$ & No \\
RPA-GP-3 & RBF & Gaussian & $3 \times 1, 3 \times 2, 3 \times 3, 2 \times 4, 2 \times 5, 1 \times 6$ & No 
\\
RPA-GP-1 & RBF & Gaussian & $20 \times 1$ & No  \\
RPA-GP-SKI & RBF & Gaussian & $20 \times 1$ & No \\
DPA-GP & RBF & Maximize Eq. (\ref{eq:objective}) & $20 \times 1$ & No \\
RPA-GP-ARD & RBF & Gaussian & $20 \times 1 $ & Yes \\
DPA-GP-ARD & RBF & Maximize Eq. (\ref{eq:objective}) & $20 \times 1 $ & Yes \\
DPA-GP-ARD-SKI & RBF & Maximize Eq. (\ref{eq:objective}) & $20 \times 1$ & Yes \\
\end{tabular} 
\caption{Summary of each evaluated model in Table \ref{table:supplemental test table}.}
\end{table*}
\renewcommand{\figwidth}{.32\textwidth}
\renewcommand{\figheight}{.25\textwidth}
\begin{figure*}[h!]
    \centering
    \subfigure{\centering \includegraphics[width=\figwidth, height=\figheight]{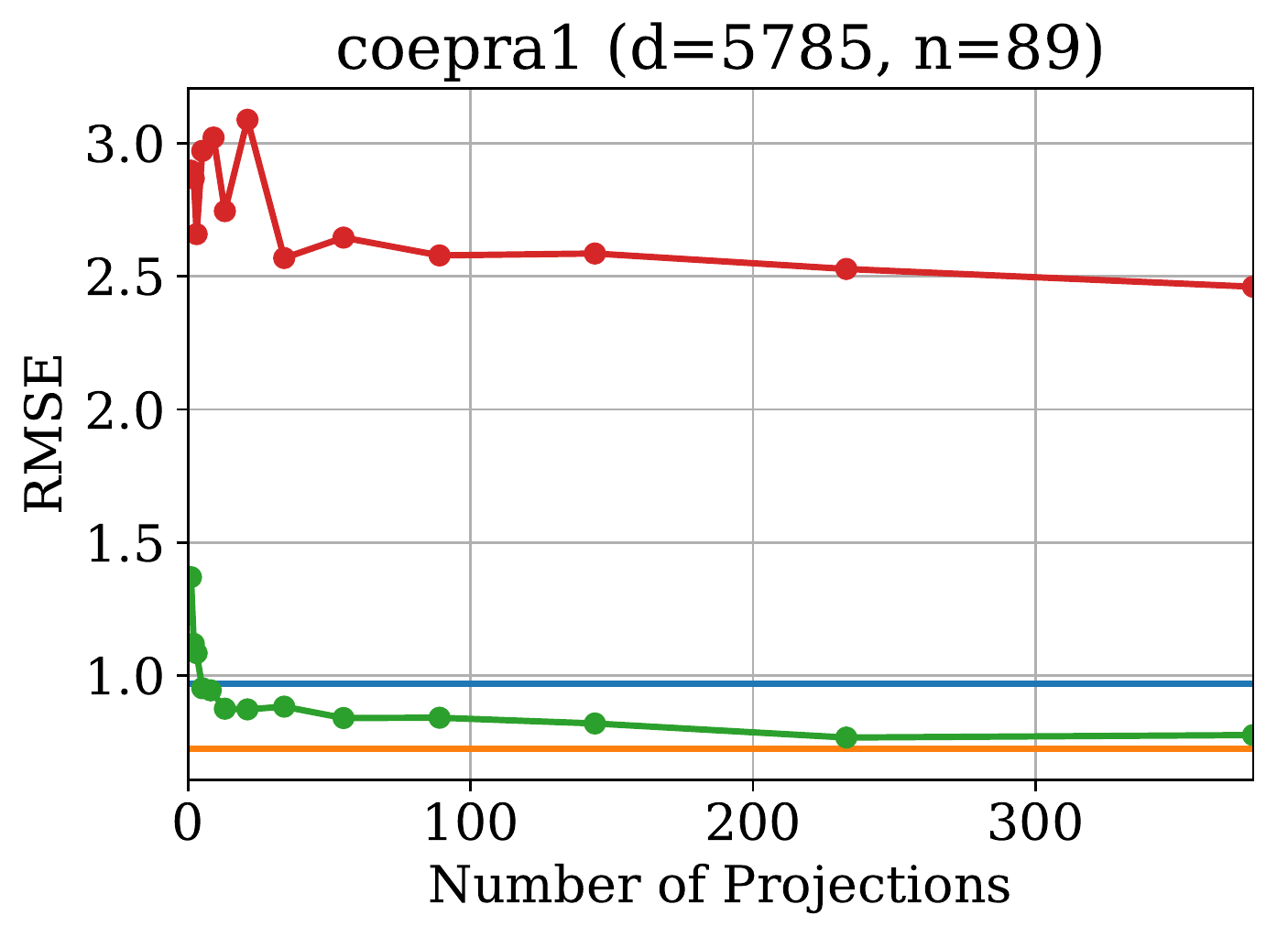}}
    \subfigure{\centering \includegraphics[width=\figwidth, height=\figheight]{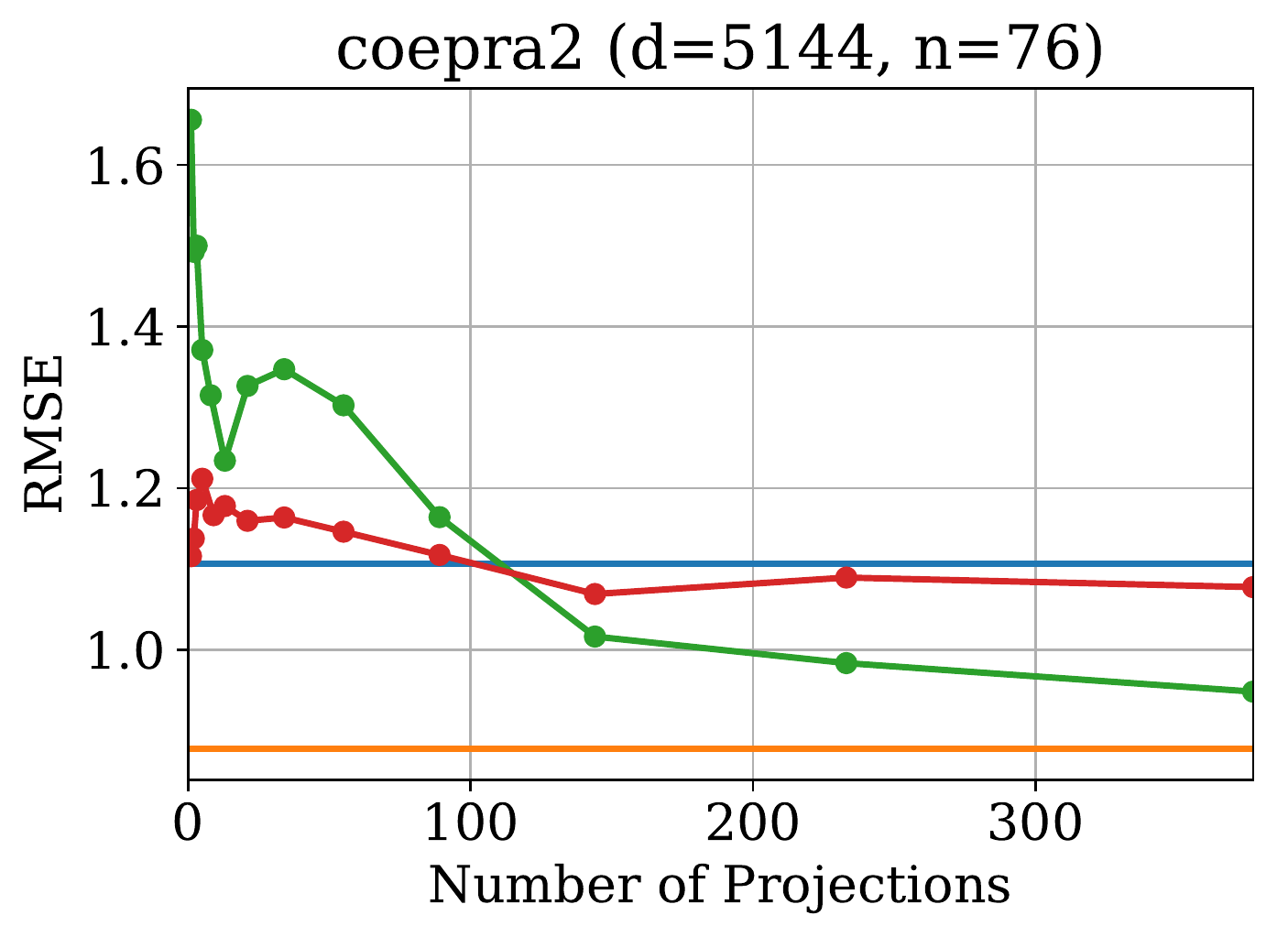}}
    \subfigure{\centering \includegraphics[width=\figwidth, height=\figheight]{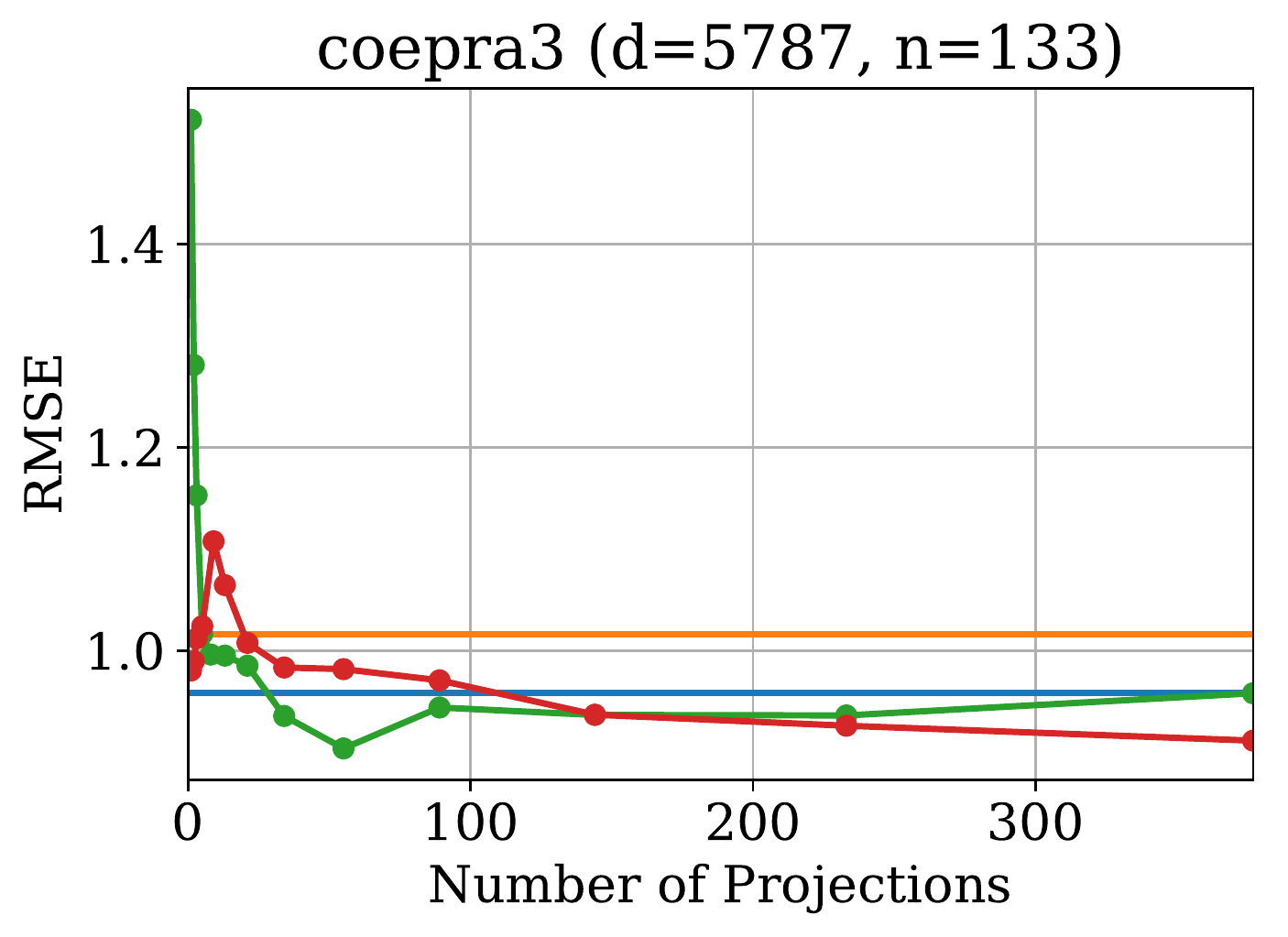}}
    \subfigure{\centering \includegraphics[width=\figwidth, height=\figheight]{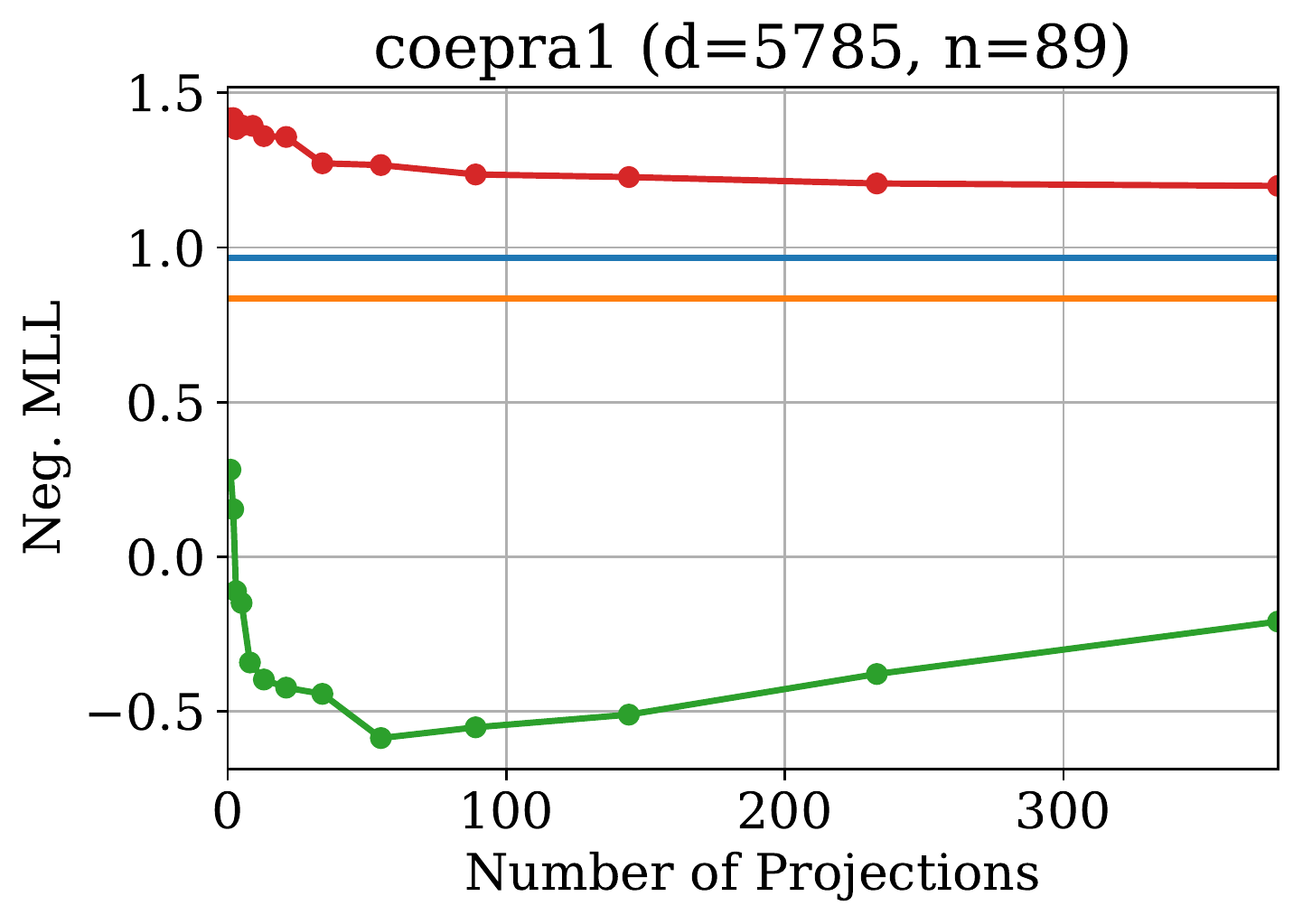}}
    \subfigure{\centering \includegraphics[width=\figwidth, height=\figheight]{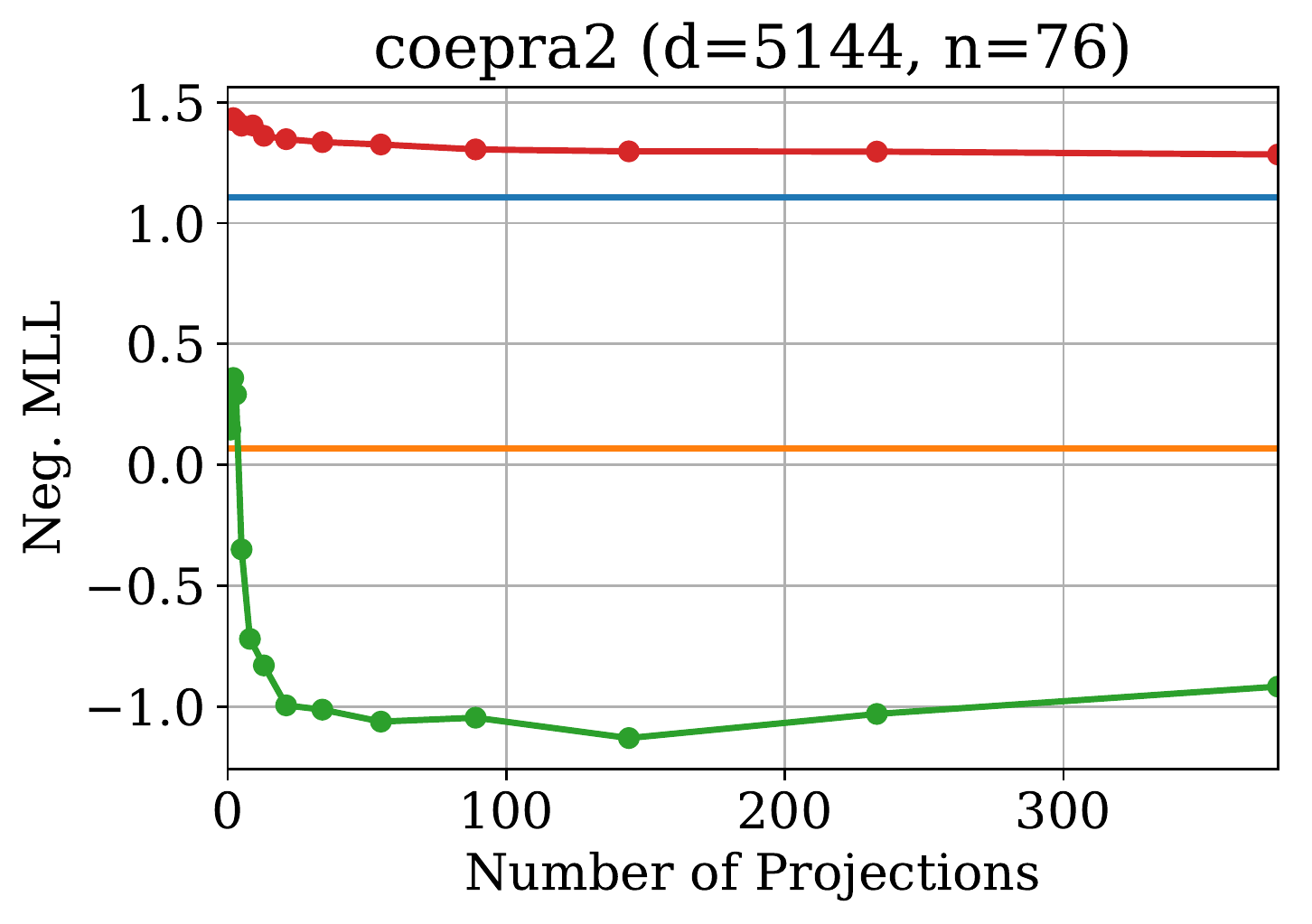}}
    \subfigure{\centering \includegraphics[width=\figwidth, height=\figheight]{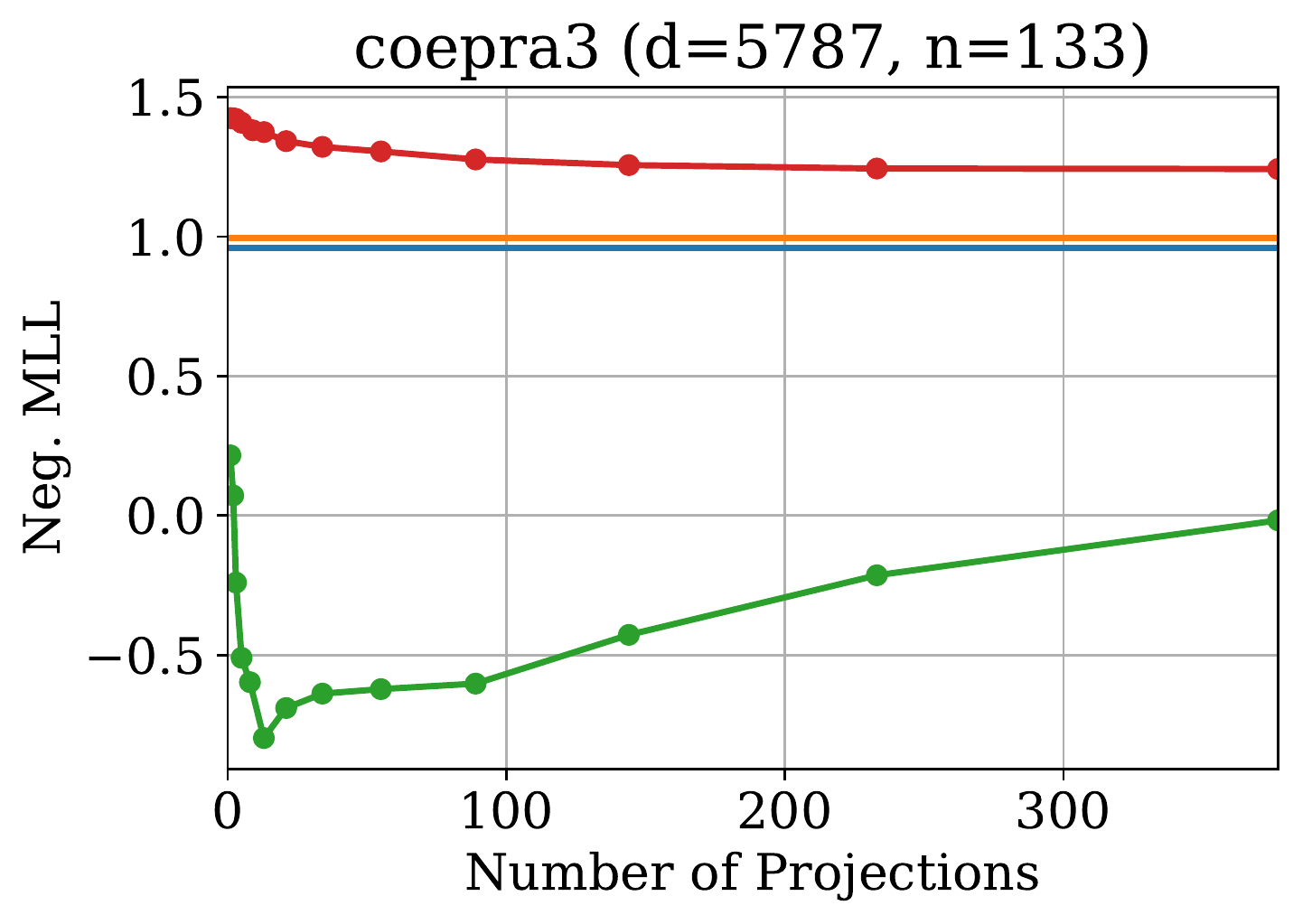}}
    \subfigure{\centering \includegraphics[width=\figwidth, height=\figheight]{imgs/orientation400.pdf}}
    \subfigure{\centering \includegraphics[width=\figwidth, height=\figheight]{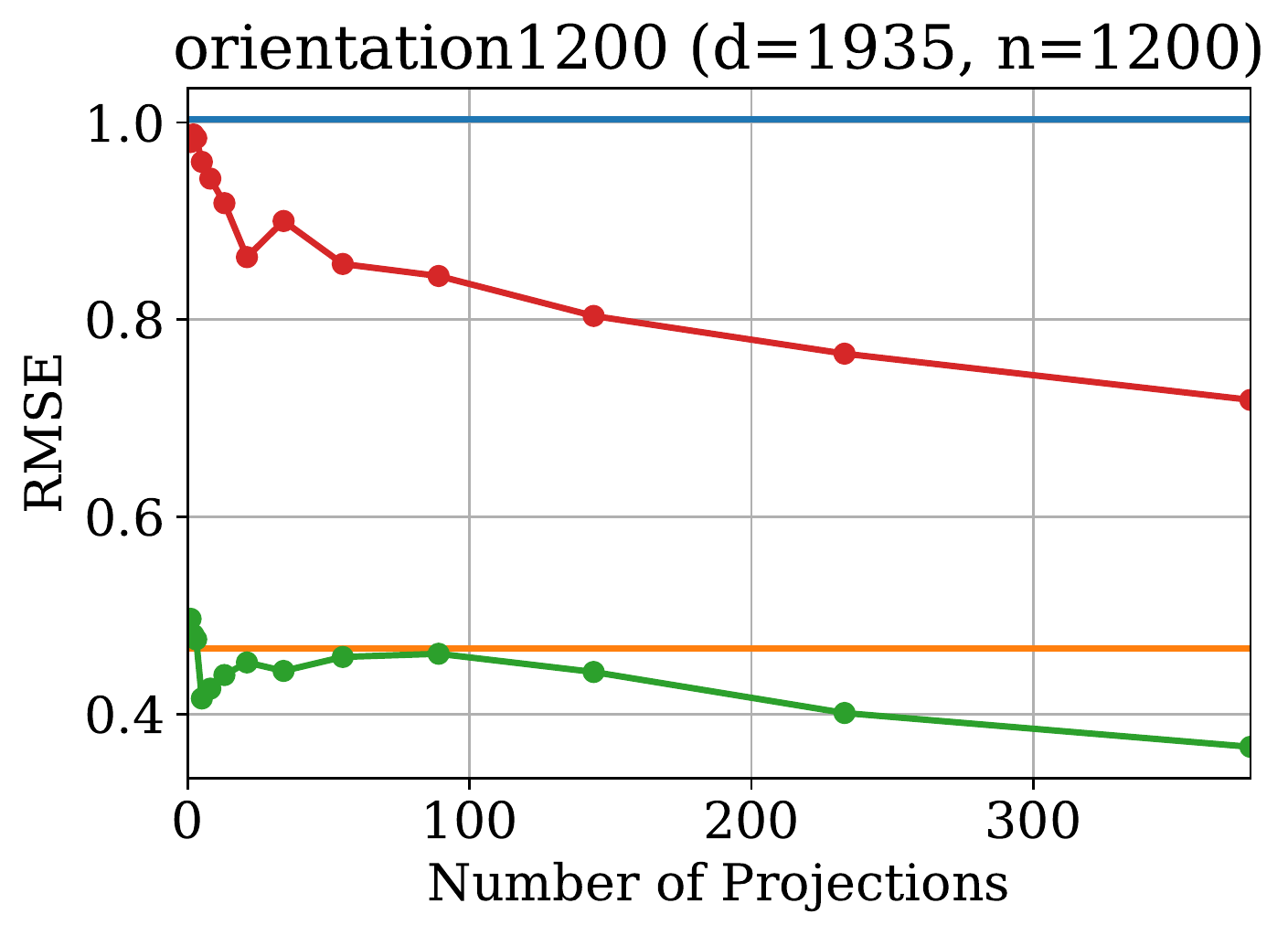}}
    \subfigure{\centering \includegraphics[width=\figwidth, height=\figheight]{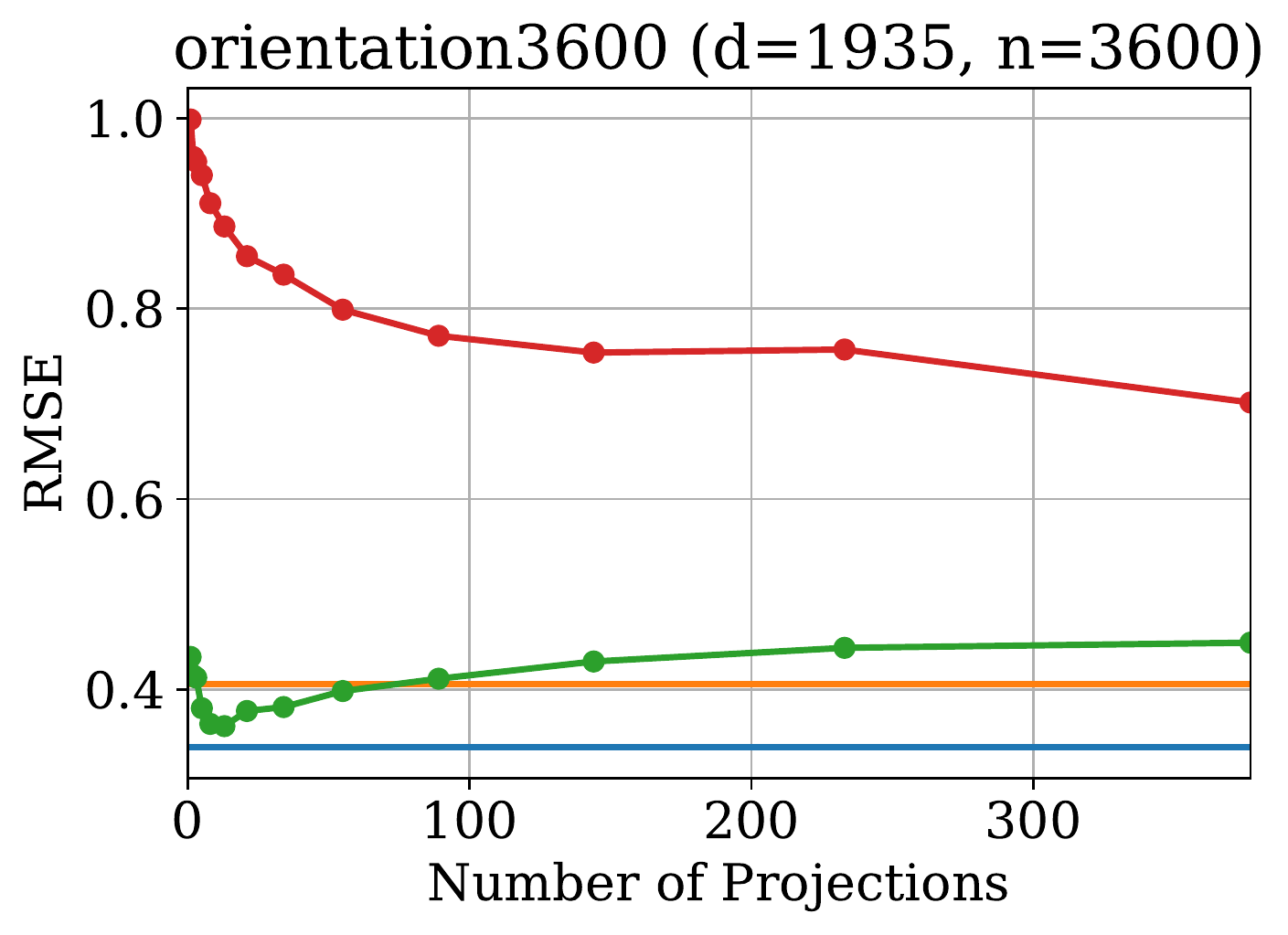}}
    \caption{\textbf{Top}: RBF-ARD GP, GAM GP, RPA-GP, and DPA-GP-ARD average RMSE on the high dimensional CoEPrA data sets. DPA-GP-ARD is competitive with GAM on these data sets for the optimal number of projections. \textbf{Middle}: training negative marginal log likelihood for the same models. For each case, marginal likelihood greatly favors DPA-GP-ARD with few projections though the optimal DPA-GP-ARD model by marginal likelihood need not be the optimal model by RMSE. \textbf{Bottom}: RBF-ARD GP, GAM GP, RPA-GP, and DPA-GP-ARD average RMSE on Olivetti face orientation data set. RBF-ARD attributes all data to noise for $n=400,1200$, and DPA-GP-ARD achieves low error with less data than either GAM or RBF-ARD.}
    \label{fig:coepra}
\end{figure*}

\section{Tests on very high-dimensional data sets}
Above in Figure \ref{fig:coepra} are the full set of plots corresponding to tests on the very high-dimensional data sets.

\end{document}